%% file: article.tex
\documentclass{elsarticle}
\usepackage{lineno, hyperref}
\modulolinenumbers[5]

\journal{Applied Soft Computing}

\usepackage{graphicx,url}
\usepackage{subfig}
\usepackage{caption}
\usepackage{amsmath}
\usepackage{amssymb}
\usepackage{booktabs}
\usepackage{multirow}
\usepackage[table,xcdraw]{xcolor}
\usepackage{longtable}

\begin{document}

\begin{frontmatter}

\title{The choice of scaling technique matters for classification performance}

\author[cinufpe,icufal]{Lucas B.V. de Amorim \corref{mycorrespondingauthor}}
\cortext[mycorrespondingauthor]{Corresponding author}

\author[cinufpe]{George D.C. Cavalcanti}

\author[ets]{Rafael M.O. Cruz}

\address[cinufpe]{Centro de Informática - Universidade Federal de Pernambuco}
\address[icufal]{Instituto de Computação - Universidade Federal de Alagoas}
\address[ets]{École de Technologie Supérieure, Université du Québec}

\begin{abstract}
Dataset scaling, also known as normalization, is an essential preprocessing step in a machine learning pipeline. It is aimed at adjusting attributes scales in a way that they all vary within the same range.
This transformation is known to improve the performance of classification models, but there are several scaling techniques to choose from, and this choice is not generally done carefully. 
In this paper, we execute a broad experiment comparing the impact of 5 scaling techniques on the performances of 20 classification algorithms among monolithic and ensemble models, applying them to 82 publicly available datasets with varying imbalance ratios. 
Results show that the choice of scaling technique matters for classification performance, and the performance difference between the best and the worst scaling technique is relevant and statistically significant in most cases. They also indicate that choosing an inadequate technique can be more detrimental to classification performance than not scaling the data at all. We also show how the performance variation of an ensemble model, considering different scaling techniques, tends to be dictated by that of its base model. Finally, we discuss the relationship between a model's sensitivity to the choice of scaling technique and its performance and provide insights into its applicability on different model deployment scenarios. Full results and source code for the experiments in this paper are available in a GitHub repository.\footnote{https://github.com/amorimlb/scaling\_matters}
\end{abstract}


\end{frontmatter}

\section{Introduction}
\label{sec:intro}

In a classification task, scaling, also called normalization, is used as an essential preprocessing step to adequate data such that every feature varies within the same range. During the model learning process, this assures that features with higher or wider numerical ranges do not dominate those that vary within a narrower or lower range, a phenomenon that may bias the analysis towards less important (less informative) attributes simply because they are on a larger scale. Dataset scaling is thus able to mitigate this phenomenon and consequently improve classification performance.

Most researchers apply this preprocessing step by default in their analysis pipelines, regardless of the classification algorithm or the dataset being used. It thus seems that the benefits promoted by dataset scaling to classification performance are common sense in the machine learning literature. Ratifying this common sense, Singh et al. \cite{singh2020a} performed a brief review of papers in different application domains that compared the performance of classifiers trained with nonscaled data to that of classifiers trained with a chosen scaling technique. Their findings confirm that scaling the data previous to model training can lead to significant performance improvements when compared to nonscaled data. 

In spite of that, we could estimate that only 16 out of the 50 most cited classification papers since 2018 using the KEEL\cite{keel_repo} dataset repository\footnote{The full list of these papers can be found in the supplemental material in the repository.} mentioned that this preprocessing step was applied. We believe the remaining 34 papers most likely applied dataset scaling, albeit not mentioning it. This shows how this crucial and ubiquitous preprocessing step is not being given the deserved attention in the literature.

Moreover, there are several scaling techniques that can be applied to a dataset and choosing the best one is by itself an important methodological decision. This decision must be carefully addressed, as it has been reported \cite{singh2020a, Mishkov2022} that choosing the wrong technique can be more detrimental to the classification performance than not scaling the data at all.

Notwithstanding, only a few studies investigate how these scaling techniques compare to each other in terms of the resulting classification performance when they are applied as a preprocessing step to different classification algorithms. Some of these studies focus their analysis on just two scaling techniques \cite{jain2018, dzierzak2019}, while those that experiment with a more extensive number of techniques restrict their tests to just one  \cite{singh2020a, Mishkov2022} or three classification algorithms \cite{raju2020}, as we detail in Section \ref{sec:related}. The motivation of this paper, therefore, lies in a combination of all these factors: the pervasiveness of scaling techniques and, at the same time, the lack of attention to it in the research community along with the shortage of empirical studies about their impact on classification performance. This is why we focus our research at answering important questions on how the performance of different classification algorithms varies when the data is scaled with distinct scaling techniques.

In this paper, we performed a broader analysis, considering different types of algorithms and also a larger number of datasets to which we apply a relevant number of scaling techniques. By broadening our choices of methods and datasets, we are able to draw conclusions that can more generalized as opposed to a more limited experiment. With that, we intend to understand if scaling affects classification performance and if the choice of scaling technique significantly causes variations in the magnitude of this influence in performance. Moreover, we want to know if these variations behave differently depending on the classification model used, including both monolithic and ensemble models. 

Ensemble models combine the outputs of monolithic classifiers to obtain a combined decision that is possibly better than the decisions of each individual classifier in this set (e.g. Random Forests, XGBoost). For the ensemble models, we included models that employ dynamic selection (Dynamic Classifier Selection and Dynamic Ensemble Selection), as these models have shown promising results when compared to monolithic ones \cite{Cruz2018survey}, which compels their addition to our investigation. To simplify our analysis we included only homogeneous ensembles, in which the base models are instances of a single classifier algorithm.

The inclusion of ensemble models in our experiments also allows us to contribute with two other important and novel analysis: (i) the relation between the rank of best choices of scaling techniques for an homogeneous ensemble and that for its base model, and (ii) the relation between the sensitivity to the choice of scaling technique and the performance of the models, which enabled us to discuss how the ensemble models can be built to manipulate this relation.

Additionally, we looked at how the observed performance variations due to the choice of scaling technique behave when dealing with data presenting different imbalance ratios (IR) since, in real-world data, it is common to have a significant imbalance between the number of instances of each class. For example, in medical diagnosis or face detection problems, when there are many more examples of the negative class than those of the positive class. At the time of writing this paper, we were unable to identify other studies that have explored the effects of scaling techniques when applied to data with different IRs.

To guide this study, we declare the following research questions: 
\begin{quote} 
    \textbf{RQ1} - Does the choice of scaling technique matters for classification performance?
\end{quote}

\begin{quote} 
    \textbf{RQ2} - Which models present greater performance variations when datasets are scaled with different techniques?
\end{quote}

\begin{quote}
    \textbf{RQ3} - Do homogeneous ensembles tend to follow the performance variation pattern presented by its base model when dealt with different scaling techniques? 
\end{quote}

In order to answer these questions and perform further analysis, we modeled an experiment using 82 datasets from the KEEL repository after preprocessing them with five different scaling techniques. We applied 20 different classification models to these datasets, including 11 monolithic and 9 ensemble models. The datasets come from different domains, with varying number of instances, features, and with IRs ranging from low to high, i.e. from virtually balanced to extremely imbalanced.  We employed 5-fold cross validation and measured classification performance according to two different metrics: F1 and G-Mean. Hypothesis tests were employed to assess the significance of the results.

To summarize, these are the main contributions of this paper: 
\begin{itemize}
    \item It empirically shows that the choice of scaling technique matters for classification performance. This is based on a broad experiment considering different types and a relevant number of classification algorithms, scaling techniques and datasets. 
    \item It is demonstrated that there is a relation between the rank of best choices of scaling techniques for an homogeneous ensemble and that for its base model. This is an important finding, since it makes it less costly to select an optimal scaling technique for an ensemble model.
    \item It demonstrates that the performance variation due to the choice of different scaling techniques tends to be more salient for datasets with higher imbalance ratios.
    \item It analyzes the relation between the sensitivity to the choice of scaling technique and the performance of the models, and discusses how the ensemble models can be built to manipulate this relation.
\end{itemize}

The rest of this paper is organized as follows:  Section \ref{sec:scaling_tech} discusses the different scaling techniques available. Section \ref{sec:class_algo} presents the classification algorithms used in this paper. The experiment methodology is covered in Section \ref{sec:method} and their results and discussions are presented in Section \ref{sec:results}. Section \ref{sec:related} reviews the related works on the comparison of scaling techniques. Finally, Section \ref{sec:lessons} summarizes the lessons learned and Section \ref{sec:conclusion} concludes this paper.

\section{Scaling techniques}
\label{sec:scaling_tech}

The raw data that generally come in real-world originated datasets often have issues that preclude them from being effectively explored in a machine learning task. One of the most prominent issues is the different scales in which the various dataset features are presented. In the case of predictive machine learning, this issue causes the algorithms to learn less accurate models. This is due to algorithms' tendency to rely on features that vary within a wider range, i.e., the dominant features, even though these are not necessarily the most informative or decisive features for the correct classification of data instances. To tackle this problem, scaling techniques are employed to adequate data such that every feature varies within the same range. Scaling is thus one of many preprocessing steps that generally have to be undertaken before applying a machine learning model to a dataset.

It is necessary to highlight that although these techniques are more commonly called normalization techniques in the machine learning literature \cite{singh2020a, Mishkov2022, jain2018, dzierzak2019, raju2020} we prefer to use the term scaling because we believe it has a more general meaning. It covers both normalization or standardization methods, in the strict statistical meaning of these words, as well as simpler scaling methods, such as the Maximum Absolute Scaler. 

In simple terms, the most common scaling techniques can be broken down into two components: a translational term and a scaling factor. Suppose one wants to transform a vector $x$, then, in Equation \ref{eq:generic} each one of its components $x_i$ is transformed into $x_i'$, where $T$ represents the translational term and $S$ the scaling factor. The translational term operates by moving data along the X-axis, while the scaling factor makes data more concentrated or spread out horizontally.

\begin{equation}
    x_i' = \frac{x_i - T}{S}
    \label{eq:generic}
\end{equation}

Some simpler techniques may not include one of these two aforementioned operations. For example, the Mean Centering technique (Equation \ref{eq:mean_cent}) simply subtracts the vector's mean from each of its components. In this case, the vector's mean is the translational term, and no actual scaling is applied.
\begin{equation}
    x_i' = x_i - \bar x
    \label{eq:mean_cent}
\end{equation}

While the Mean Centering effectively removes the offset from the data, shifting its mean to zero, it fails to equalize data variances across the different features of a dataset. More elaborated techniques promote this equalization by multiplying the data by a scaling factor such that it varies across a defined range, e.g. [-1, 1] or [0,1].

The following subsections present each of the five scaling techniques used in this study: Standard Scaler, Min-max Scaler, Maximum Absolute Scaler, Robust Scaler and Quantile Transformer. These are five well known, diverse, and commonly used scaling techniques. We believe these techniques are good representatives of the most relevant techniques used by the machine learning community. According to the literature, these techniques are diverse and non redundant  \cite{singh2020a, Mishkov2022}. 

\begin{figure*}[ht!]
    \centering
    \subfloat[\label{fig:comparing_STs_pos_neg}]{
        \includegraphics[trim=5 5 5 5,clip,
        height=16.1cm]{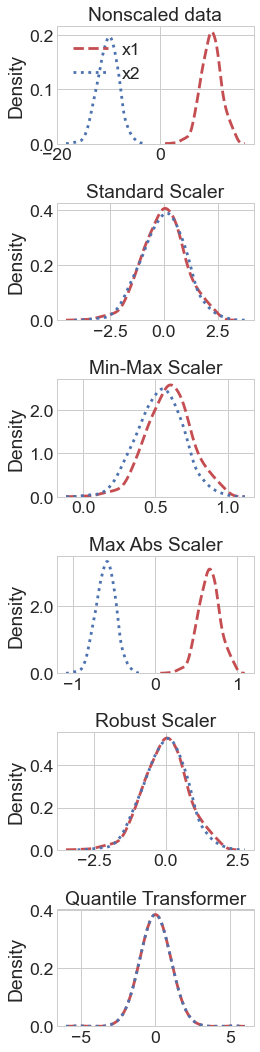}
    }
    \subfloat[\label{fig:comparing_STs_outliers}]{
        \includegraphics[trim=32 5 5 5,clip,
        height=16.1cm]{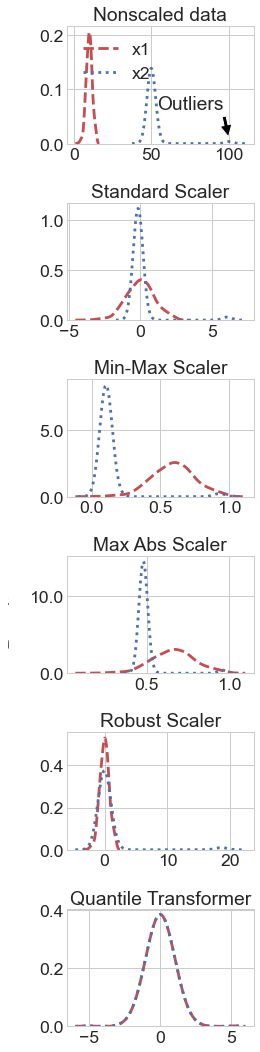}
    }
    \subfloat[\label{fig:comparing_STs_normal_uniform}]{
        \includegraphics[trim=30 5 5 5,clip,
        height=16.1cm]{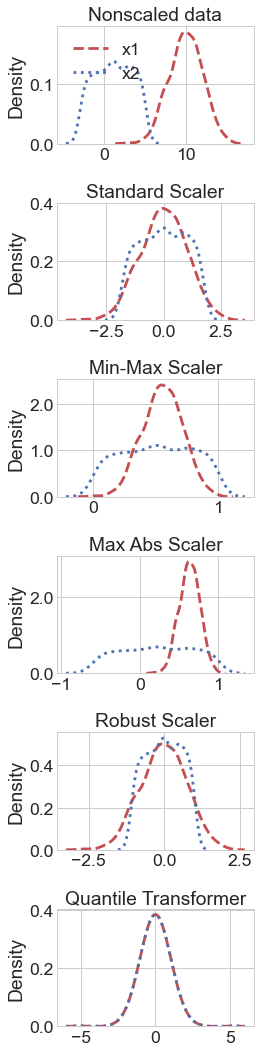}
    }
    \caption{Effects of the selected scaling techniques when given as input: (a) normally distributed variables with opposing means and no outliers, (b) normally distributed variables where $x_2$ presents outliers and (c) variables with different distribution shapes: $x_1$ has a normal distribution and $x_2$ has a uniform distribution. 
    }
    \label{fig:comparing_STs}
\end{figure*}

Figure \ref{fig:comparing_STs} presents examples where these techniques scale pairs of attributes, providing a visual feedback as to how these techniques work. Each column of the figure corresponds to one example. The original data distribution is always shown in the topmost graph. In Figure \ref{fig:comparing_STs_pos_neg}, both variables were randomly generated to follow normal distributions with chosen means and variances. Precisely, if we represent our randomly generated normal variable as $x_i \sim \mathcal{N}(\mu, \sigma^2, n) $, where $\mu$ is its mean, $\sigma^2$ its variance, and $n$ is the sample size, then: $x_1 \sim \mathcal{N}(10, 4, 1000)$ and $x_2 \sim \mathcal{N}(-10, 4, 1000)$. Here the intent is to see how the scaling techniques deal with variables in opposite sides of the X-axis.

In Figure \ref{fig:comparing_STs_outliers}, the two variables were also randomly generated to follow normal distributions, but 25 outliers are added to $x_2$ near the 100 value. Precisely: $x_1 \sim \mathcal{N}(10, 4, 1000)$ and $x_2 \sim \mathcal{N}(50, 4, 975) + \mathcal{N}(100, 4, 25)$. This figure shows how the effects of the scaling techniques on a variable that presents outliers compares to their effects on a variable without outliers.

Finally, Figure \ref{fig:comparing_STs_normal_uniform} shows the effects of the techniques in attributes with different distribution shapes: While $x_1$ is, similar to the previous figures ($x_1 \sim \mathcal{N}(10, 4, 1000)$), $x_2$ is a randomly generated sample from a uniform distribution in the [-3, 5] range, with a sample size of 1000. The goal of this figure is to show how each technique manipulates the distributions' shapes. 


\subsection{Standard Scaler}

The Standard Scaler technique, which implements the Z-score normalization, standardizes attributes by subtracting their mean from each value and dividing the result by the attribute's standard deviation $s$, resulting in a distribution with zero mean and unit variance. Let $\bar x$ be the mean of the $x$ variable, a value $x_i$ is transformed (scaled) into $x_i'$ by means of Equation \ref{eq:ss}.

\begin{equation}
    x_i' = \frac{x_i - \bar x}{s}
    \label{eq:ss}
\end{equation}

In this case, the translational term is the attribute's sample mean, while the scaling factor is its standard deviation. One advantage of this technique is that it can transform both positive and negative valued attributes into a very similar distribution, as can be seen in Figure \ref{fig:comparing_STs_pos_neg}. However, in the presence of outliers, it makes the final distribution of inliers too narrow when compared to that of an attribute without outliers (Figure \ref{fig:comparing_STs_outliers}). 

Variations of this technique are the Pareto Scaling \cite{eriksson1999} (Eq. \ref{eq:pareto}) and Variable Stability (Vast) scaling \cite{Keun2003} (Eq. \ref{eq:vast}). They alter the scaling factor, making the resulting distribution wider. This way, they reduce the importance of the outliers but are not entirely immune to them either.

\begin{equation}
    x_i' = \frac{x_i - \bar x}{\sqrt{s}}
    \label{eq:pareto}
\end{equation}

\begin{equation}
    x_i' = \frac{x_i - \bar x}{s} \frac{\bar x}{s}
    \label{eq:vast}
\end{equation}

\subsection{Min-max Scaler}

The Min-max Scaler alters an attribute scale and shifts its values along the X axis so that the transformed attribute ranges within the $[0, 1]$ interval, according to Equation \ref{eq:mms}.

\begin{equation}
    x_i' = \frac{x_i - x_{\text{min}}}{x_{\text{max}} - x_{\text{min}}}
    \label{eq:mms}
\end{equation}

In this technique, the scaling factor consists of the attribute's range, and the translational term is its minimum value. This way, this technique ensures a new minimum of zero and a new maximum of one. For attributes that do not present outliers, the Min-max Scaler has an effect similar to the Standard Scaler. Albeit, in the case of the latter, the resulting distribution  will be in a less strictly defined range, as can be observed in Figure \ref{fig:comparing_STs_pos_neg}. On the other hand, when data present outliers, this technique fails to equalize both the means and variances of the distributions (see Figure \ref{fig:comparing_STs_outliers}) and is thus generally unsuitable for a machine learning pipeline.

The Min-max scaler can be generalized in Equation \ref{eq:mms_gen} in order to allow for the definition of the resulting range as [a, b] instead of [0, 1]. In fact, we notice that machine learning researchers frequently use this second form to achieve a [-1, 1] range.

\begin{equation}
    x_i' = a + \frac{(x_i - x_{\text{min}})(b - a)}{x_{\text{max}} - x_{\text{min}}}
    \label{eq:mms_gen}
\end{equation}

\subsection{Maximum Absolute Scaler}

The Maximum Absolute Scaler modifies the scale of an attribute by simply dividing each example by the attribute's maximum absolute value, as in Equation  \ref{eq:mas}. 

\begin{equation}
    x_i' = \frac{x_i}{\text{max}(|x|)}
    \label{eq:mas}
\end{equation}

As such, this technique is sensitive to outliers, as one can see in Figure \ref{fig:comparing_STs_outliers}, where $x_2$ is transformed into a much narrower distribution than $x_1$. Also, we notice in Figure \ref{fig:comparing_STs_pos_neg} that, since this scaling technique misses a translational term, it is unable to equalize the attribute's means. 

\subsection{Robust Scaler}

The previous three scaling techniques are very sensitive to the presence of outliers since these transformations depend on the mean or on the minimum and maximum values of each variable. The Robust Scaler seeks to mitigate the effects of outliers by centering data around the median (second quartile of $x$, $Q_2(x)$) and by scaling it according to the interquartile range, which is the magnitude of the difference between the first quartile $Q_1(x)$ and the third quartile $Q_3(x)$ of $x$, as shown in Equation \ref{eq:rs}.

\begin{equation}
    x_i' = \frac{x_i - Q_2(x)}{Q_3(x) - Q_1(x)}
    \label{eq:rs}
\end{equation}

Figure \ref{fig:comparing_STs_outliers} shows that this technique is effective in equalizing the variances among the different attributes even when one of them presents outliers. This happens because when the interquartile range is used as the scaling factor, the presence of the outliers is disregarded, since they lie beyond this interval.  

\subsection{Quantile Transformer}

This scaling technique is part of a family of techniques that perform a non-linear transformation on the data as opposed to the previously discussed techniques. It is able to change the shape of the original attribute distribution, which can be particularly useful as there has been reported evidence that transforming attributes so that they follow a Gaussian-like distribution may benefit classification performance \cite{Hu2021}. The Quantile Transformer technique transforms an attribute using its quantiles information (employing a quantile function) and puts all attributes in the same desired distribution, uniform or normal. In Figure \ref{fig:comparing_STs} we set the output distribution to `normal'.  It is also a robust technique because it is not sensitive to outliers.

This transformation is applied to each attribute independently in the following way: First, the attribute's cumulative distribution function is estimated and used to map the original values to a uniform distribution. Then, the obtained values are mapped to the desired output distribution using the associated quantile function. New values that fall outside the output range are mapped to the bounds of the output distribution \cite{scikit_quantile}. 

Notice in Figure \ref{fig:comparing_STs} that the Quantile Transformer is the only scaling technique that was able to provide consistently similar resulting distributions, regardless of the original attributes given as inputs. It can be seen in Figure \ref{fig:comparing_STs_normal_uniform} that this technique was able to adequate the $x_2$ attribute, changing its distribution so that it matches a standard normal distribution, with zero mean and unit variance. This promotes better comparability of attributes that previously had different scales and distribution shapes. Although, being a non-linear transformation, it may distort linear correlations between variables measured at the same scale.

\section{Classification Algorithms}
\label{sec:class_algo}

In order to evaluate the scaling techniques over a broad choice of classification algorithms, we selected both monolithic and ensemble models spanning eight different subcategories. The complete list, along with the relevant model's parameters, is presented in Table \ref{tab:params}. Our goal is to perform an analysis whose results may be generalized for a wide range of classification algorithms. 

With that in mind, we selected monolithic models from the following subcategories: Instance-based, probabilistic, discriminant analysis, rule-based and neural networks. For the ensemble models, the algorithms are distributed among these tree subcategories: Static, DCS, DES. We chose to include ensemble models in our analysis because, (i) these methods, specially DCS and DES, have shown to be more promising than monolithic models in various scenarios \cite{Cruz2018survey}, (ii) we assume that, due to the effect of the combination of various classifiers, these techniques may be less sensitive to changes in the data than monolithic models. Additionally, we could not find in the literature any analysis on how these algorithms behave when dealt with different scaling techniques.

\input{tabs/tab_params}

\subsection{Monolithic Models}

\subsubsection{Instance-based algorithms}
Instance-based algorithms rely on a specific subset of instances, rather than a model representing the whole set, to classify each query instance. For example, in the case of Nearest Neighbors classifiers \cite{Cover1967}, the algorithm uses the training data points closest to a query instance in the feature space to determine its class. 

This paper includes the following instance-based algorithms in the experiments: the $k$-Nearest Neighbors (KNN), GLVQ \cite{glvq} and SVM \cite{svm}. For all of these algorithms, we used the implementation provided by the Scikit-learn Python library (sklearn) \cite{sklearn}. 

The KNN algorithm classifies a new (query) instance by first consulting the labels of $k$ training instances that are nearest to the query instance, and then assigning the most common label to the query instance. The GLVQ algorithm, which is a generalization of the LVQ algorithm, first learns prototypes to represent the instances and then classify query instances based on their distances to the prototypes. The SVM algorithm tries to maximize the distances between instances of different classes by creating a hyperplane that divides the classes while being equidistant to the most ``marginal'' instances of both classes, the support vectors. A query instance is assigned to a class according to its position in relation to the hyperplane. Note that one can say that GLVQ and SVM are not purely instance-based, since both of them learn an abstraction of the instances.

\subsubsection{Probabilistic algorithms}
Probabilistic algorithms rely on probability models to calculate the probability that a specific instance belongs to a class. Naive Bayes classifiers well represent this category. Naive Bayes uses Bayes' theorem with the ``naive'' assumption of conditional independence between the pairs of attributes given a particular class. Despite this simplification, they have proven useful classifiers even when this assumption does not hold \cite{Rish2001}. Although they are not precise probability estimators in this case \cite{Zhang2004}. An implementation of such classifiers is the Gaussian Naive Bayes (GNB), it assumes the likelihood of features as a Gaussian distribution. This paper includes the GNB and another type of probabilistic classifier, the Gaussian Process (GP) classifier \cite{Seeger2004}.

\subsubsection{Discriminant Analysis}

Discriminant analysis is a family of methods that aim at finding decision surfaces that are most optimally able to separate the classes of the data instances in the feature space. Linear Discriminant Analysis (LDA) and Quadratic Discriminant Analysis (QDA) are both included in our experiments. In LDA, the algorithm seeks linear decision surfaces, while in QDA, it seeks quadratic decision surfaces. These methods are also commonly used for dimensionality reduction. 

\subsubsection{Rule-based algorithms}

Rule-based classifiers are those that make use of IF-THEN rules for class prediction \cite{Tung2009}. In this way, we may consider Decision Tree as a rule-based model because it learns these rules, with which it builds the tree. In this paper, we have included the Decision Tree (DT) classifier included in sklearn that implements an optimized version of the Classification and Regression Trees (CART) algorithm \cite{Breiman20171}.

\subsubsection{Neural Networks}

Artificial Neural Networks (ANNs), which are also referred to as simply ``Neural Networks'', are algorithms inspired by the mechanisms present in the biological neural networks in our brains. ANNs contain computational units, called neurons, connected through weights, which mimic the role of the strengths of synaptic connections in biological neural networks \cite{Aggarwal2018}. 

The simplest ANN architecture is the Single Layer Perceptron, or simply Perceptron. It contains a single input layer and one output node. Similarly to LDA, it looks for a linear decision surface that better separates the classes of a problem. Multilayer Perceptrons are more complex models that contain multiple computational layers. Each layer feeds its outputs to the inputs of the next layer, which allows for more complex computations. The resulting decision surfaces are nonlinear and hence more flexible, to the extent that may even lead to overfitting, which may be addressed with a better tuning of the model's hyperparameters. This paper includes both a single layer (Percep) and a multilayer perceptron (MLP).

\subsection{Ensemble Models}

Ensemble models, also called Multiple Classifier Systems or Committees of Classifiers, integrate the decisions of a set (ensemble) of classifiers aiming to obtain a combined decision that is better than the decisions of each individual classifier in the ensemble. Theoretical and empirical studies have shown that an ensemble is typically more accurate than an individual classifier \cite{Kuncheva2014}.

The classifiers inside an ensemble are called base classifiers. Most ensemble models use a single classifier algorithm to generate all the base classifiers. These ensembles are called homogeneous ensembles. On the other hand, when the base classifiers are generated by multiple methods, the ensemble is called a heterogeneous ensemble. For the sake of simplicity, we included only homogeneous ensembles in this work.  

When it comes to how the multiple base classifiers' decisions are combined, there are roughly two categories: static and dynamic ensembles. Static ensembles usually generate their decisions by combining the base classifier outputs through a majority voting rule, where the most frequent decision is chosen, or a variation of this mechanism such as the weighted majority voting \cite{Zhou2012, Tulyakov2008}. The most widely known and employed ensemble models fall within the static ensemble category.

Dynamic ensembles select either a single classifier (DCS - Dynamic Classifier Selection), or a subset of the original ensemble (DES - Dynamic Ensemble Selection) to label a query instance. In both cases, the models consider the local region where the query instance lies, defined by its $k$ nearest training instances, and try to select the best performing base classifier(s) in that region. This local region is called the Region of Competence (RoC) of the input query. 

Dynamic ensembles have shown to be very promising for many different scenarios \cite{Cruz2018survey}, making these models a compelling addition to our arsenal. Moreover, there are no previous studies on how their performances vary under different scaling techniques, which is an interesting investigation since, in their selection phase, these methods use this notion of a local region of competence which is prone to changes due to dataset scaling. This paper includes nine ensemble models, which we describe below. As can be seen in Table \ref{tab:params}, for the ensemble models, we used the implementations provided by the Scikit-learn \cite{sklearn}, XGBoost \cite{Chen2016} and the DESlib \cite{DESlib} Python libraries.

\subsubsection{Static Ensembles}

Four static ensembles representing the most common approaches are considered in this paper.

\begin{description}
    \item[Bagging] Bagging, an acronym to \textit{Bootstrap AGGregatING} was introduced by Leo Breiman, in 1996 \cite{Breiman96}. Its idea is to create an ensemble comprising diverse base classifiers by giving them bootstrap replicates of the training set \cite{Kuncheva2014}. In other words, each base classifier is trained on a different subset of the labeled data, usually generated by random sampling with replacement.  The outputs of the various base classifiers are combined with majority voting.
    
    \item[Random Forests] Also introduced by Leo Breiman, in 2001 \cite{Breiman2001}, Random Forests is an extension of Bagging, albeit specifically designed for decision tree ensembles, where the major difference is that it incorporates randomized feature selection \cite{Kuncheva2014}. When building each decision tree in the ensemble, besides giving them a bootstrap copy of the training set, the algorithm performs the conventional decision tree split procedure but within a random subset of the features.
    
    \item[AdaBoost] AdaBoost, whose name derives from \textit{ADAptive BOOSTing}, is a boosting technique introduced by Freund and Schapire in 1997 \cite{Freund97}. This technique works by iteratively adding one classifier at a time. Each classifier is trained on a selectively sampled subset of the training instances, where the sampling distribution starts uniform but changes with each iteration by attributing higher selection probabilities to the instances that were misclassified in the previous iteration \cite{Kuncheva2014}. This way, the algorithm gathers classifiers that have their training each time more focused on the harder instances of the training set.
    
    \item[XGBoost] XGBoost provides optimizations over the original  GBM (Gradient Boosting Machine) algorithm \cite{Friedman2001} aiming to make it more scalable and accurate. Gradient Boosting Machine, or Gradient Tree Boosting, in its turn can be understood as a generalization of AdaBoost. It also gradually adds classifiers (in this case, Decision Trees) to the ensemble. The main difference lies in how it boosts the base classifiers: instead of assigning higher weights to harder instances, GBM minimizes a differentiable loss function that can be user-defined. A tree is added to the ensemble if it reduces the loss. XGBoost adds regularization to control overfitting, parallel processing, ways to deal with sparse data, and other modifications that improve its performance over previous boosting methods \cite{Chen2016}.

\end{description}

\subsubsection{Dynamic Ensembles}

Five dynamic ensembles are included in our experiments: 3 DCS and 2 DES strategies. 

\begin{description}
    \item[Overall Local Accuracy (OLA)] OLA is a simple DCS method that evaluates the competence (accuracy) of each base classifier in the RoC of the query instance and then selects the one that presents the highest accuracy. The output of the selected classifier determines the ensemble's decision \cite{Woods97}.
    
    \item[Local Class Accuracy (LCA)] LCA is a DCS method that evaluates the local competence of each base classifier with regards to a specific class. Given that a base classifier predicts class $w_l$ for the query instance, its competence is the percentage of training samples in the RoC labeled as $w_l$ that it correctly predicts as $w_l$. The most competent classifier then determines the output of the ensemble \cite{Woods97}. 
    
    \item[Multiple Classifier Behavior (MCB)] MCB is a DCS method where the RoC is defined both by the kNN approach and by the behavioral knowledge (BKS) method \cite{Giacinto2001}. First, the outputs of all base classifiers are calculated for all training instances. Then a initial RoC is defined by selecting the $k$ nearest neighbors. This RoC is then filtered by selecting only the instances that are similar enough to the query instance based on the BKS method. The BKS method measures the similarity between the decisions of multiple classifiers when given as input a pair of instances. When this similarity is higher than a certain threshold, the instance is selected to integrate the RoC. After defining the RoC, the competence level of a classifier is calculated according to OLA. At the end, the most competent base classifier determines the ensemble output if it is better than the other classifiers by a certain threshold, else a majority voting rule is applied \cite{Britto2014}.

    \item[k-Nearest Oracles Eliminate (KNORA-E)] KNORA-E is part of the KNORA family of DES methods proposed by Ko et al. \cite{Ko2008} that is inspired by the Oracle concept \cite{Kuncheva2002}. An Oracle is an abstract classifier selection method that always selects the classifiers that return the correct output if such classifiers exist.

    For the KNORA-E method, a classifier is considered competent in a certain RoC if it achieves perfect performance for the instances within this region. Such a classifier is known as a local oracle. All local oracles are selected. In the case that no local oracles exist in the RoC, the number of neighbors that compose the RoC is iteratively reduced, by removing the farthest neighbor, until at least one classifier achieves perfect performance. When no local oracles are found, the whole pool (initial ensemble) is used for classification. In both cases, the outputs of the resulting ensemble are combined using the majority voting rule.
    
    \item[k-Nearest Oracles Union (KNORA-U)] The KNORA-U method, in its turn, selects all classifiers that produce the correct output for at least one instance in the RoC. For each instance correctly classified in the RoC, the base classifier can submit one vote to classify the query instance. The votes of all the selected classifiers are combined to produce the output, which will be the class with more votes.
    
\end{description}

\section{Experiment Methodology}
\label{sec:method}

For this experiment, we applied the five selected scaling techniques to each of the 82 original datasets selected for the study, creating 5 additional variants of each dataset. Then, we applied 20 different classification models to those datasets. In order to keep things simple, we decided not to balance the datasets, leaving them with their original class imbalance ratios. Classification performance was measured according to two different metrics: F1 and G-Mean. We chose these metrics because they consider the imbalanced nature of the datasets \cite{Akosa2017}. The selection of the scaling techniques and classification algorithms was performed prioritizing the diversity and the popularity of the methods within the machine learning community. In the case of the scaling techniques, we avoided including redundant techniques, as in previous works that used a larger number of techniques it was reported that many of them yielded similar results for most datasets \cite{singh2020a, Mishkov2022}.

The source code for the experiment, all the datasets used, and the full results table are available at the GitHub repository mentioned earlier.

\subsection{Datasets selection}

In order to allow the reproducibility of this study, we used multiple publicly available datasets, showing a wide range of IR values obtained from the KEEL dataset repository \cite{keel_repo}. These real world datasets were originally published in the UCI Machine Learning repository, which contains a collection of datasets that are frequently used by machine learning researchers for the empirical analysis of algorithms \cite{uci_repo}. The UCI repository also contains artificial datasets and data generators, but it was because of its wide range selection of real world datasets that it became one of the most popular sources for empirical machine learning studies.

The datasets we used were preprocessed by the KEEL team in a way that the multi-class problems were transformed into binary ones and also were split into files aiming its use with a 5-fold stratified cross-validation (which we employed). This way, since datasets are already pre-split, any researcher that uses this data will use the exact same folds. This also promotes research reproducibility.

First, we selected all the 91 datasets available at KEEL that are both binary and with varying imbalance ratios. Then, we selected those with a maximum 30\% categorical attributes (such that these attributes do not exert a significant influence on the results, as we are interested in the effects of scaling techniques and these only apply to numerical data), resulting in 82 datasets, as described in Table \ref{tab:descDS}. This table shows the datasets' names alongside the numbers of numerical and categorical attributes, the class counts (i.e., how many instances pertain to each class), and, finally, the imbalance ratio (IR), which is the proportion of the number of instances of the majority class to those of the minority class. The table is sorted according to the imbalance ratio.

\input{tabs/tab_descDS2}

The datasets represent problems from a diverse range of domains. There are datasets on glass identification (glass), medical analyses (pima, wisconsin, haberman, cleveland, dermatology), plant identification (iris), speech recognition (vowel), molecular and cellular biology (yeast, ecoli), image recognition (vehicle, led7digit), aeronautics (shuttle) and others.

\subsection{Datasets Scaling}

Before applying the scaling techniques, two other common preprocessing tasks were performed: strings cleaning (stripping and standardization) and one-hot encoding, which were applied to the categorical attributes. In the strings cleaning step, all the class names and values were standardized within each dataset, removing extra white spaces and correcting typos. The same was applied to the values of categorical attributes, to avoid recognizing misspelled values as new categories. In the one-hot encoding step, each categorical attribute is replaced by $n-1$ binary columns where $n$ is the number of unique values of the attribute. This way, the combination of the bits in the $n-1$ binary columns numerically encode the $n$ possible categorical values of the attribute. This ensures that the dataset can be treated by algorithms that can not deal with nonnumerical attributes.

As for the data scaling, which is the main independent variable in this study, we implemented the following procedure: We created five copies of each of the 82 original datasets, then we applied to each of the five copies one of the selected scaling techniques: Standard Scaler, Min-max Scaler, Maximum Absolute Scaler, Robust Scaler, and Quantile Transformer. We also kept an unaltered copy of the datasets (nonscaled) to serve as a baseline.

It is important to register that, in order to avoid leaking information from the test set into the training phase, a problem known as the \textit{look-ahead bias}, each of the five folds in a dataset were independently scaled according to the parameters obtained from the training set pertaining to that fold. For example, for the Min-Max technique, for every fold, the minimum and maximum values are estimated using the training set and are later used to transform the test set. 

\subsection{Performance metrics definitions}
\label{sec:metrics}

\begin{figure*}[!ht]
    \centering
    \includegraphics[trim=0 20 0 0,clip,  width=0.3\linewidth]{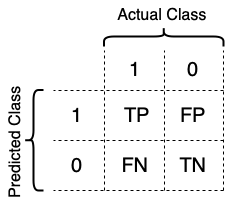}
    \caption{Confusion matrix for a two-class problem.}
    \label{fig:conf_matrix}
\end{figure*}

Classification accuracy is one of the most obvious measures of classification performance. It is calculated by the number of correctly classified examples over the total number of instances. However, this metric is not a good choice for highly imbalanced data, as it is easy to get an overly optimistic accuracy simply by classifying all instances of the test set as being of the majority class. Therefore, a different choice of metric is adamant for meaningful performance evaluation when dealing with datasets presenting different degrees of class imbalance. 
In this sense, we chose to evaluate models utilizing two metrics that consider the imbalanced nature of the datasets \cite{Akosa2017}: F1 and G-Mean. Both metrics can be defined based on the elements of a confusion matrix, as in Figure \ref{fig:conf_matrix}. 

In this matrix, TP (True Positives) and TN (True Negatives) represent the numbers of instances that were correctly classified as positive and negative, respectively. In contrast, FP (False Positives) and FN (False Negatives) represent the number of those incorrectly classified as positive and negative, respectively. In the following subsections, we present a definition for both F1 and G-Mean metrics.

\subsubsection{F1 definition}

F1, or F-score, is an harmonic mean between precision and recall (or sensitivity), which in turn are defined as in Equations \ref{eq:precision} and \ref{eq:recall}.

\begin{equation}
    \text{Precision} = \frac{TP}{TP+FP}
    \label{eq:precision}
\end{equation}

\begin{equation}
    \text{Recall} = \frac{TP}{TP+FN}
    \label{eq:recall}
\end{equation}

Then, F1 can be defined as in Equation \ref{eq:f1}.

\begin{equation}
    \text{F1} = 2\cdot \frac{\text{Precision}\cdot \text{Recall}}{\text{Precision}+\text{Recall}} = \frac{TP}{TP + \frac{1}{2}(FP+FN)}
    \label{eq:f1}
\end{equation}

F1 can be seen as a particular case of the  F$_\beta$ metric (Eq.\ref{eq:Fbeta}), with $\beta = 1$. 
\begin{equation}
    \text{F}_\beta = \frac{(1+\beta^2)\cdot TP}{(1+\beta^2)\cdot TP + \beta^2 \cdot (FP+FN)}
    \label{eq:Fbeta}
\end{equation}

\subsubsection{G-Mean definition}

G-Mean is calculated as the square root of the product of sensitivity and specificity, as in Equation \ref{eq:gmean}. Where sensitivity, or recall, can be seen as the True Positive Rate while specificity can be seen as the True Negative Rate.

\begin{equation}
    \text{Sensitivity} = \text{TPR} =  \frac{TP}{TP+FN}
\end{equation}

\begin{equation}
    \text{Specificity} = \text{TNR} = \frac{TN}{FP+TN}
\end{equation}

\begin{equation}
    \text{G-Mean} = \sqrt{\text{TPR}\cdot\text{TNR}} = \frac{\sqrt{TP \times TN}}{\sqrt{(TP+FN)(TN+FP)}}
    \label{eq:gmean}
\end{equation}

\subsection{Software}

This experiment was executed using the Python programming language, version 3.8.3 using, mostly, the following libraries: Scikit-Learn\cite{sklearn} version 0.23.1, sklearn\_lvq version 1.1.0,  DESlib\cite{DESlib} version 0.3.5 and SciPy\cite{scipy} version 1.5.0.

\section{Results and Discussion}
\label{sec:results}

This section presents graphs and tables that summarize our experimental results aiming to answer the research questions defined in Section \ref{sec:intro} and also provide further analysis.

In Section \ref{sec:answ_rq1_rq2}, in order to answer RQ1, we compare the performances obtained by models considering the different scaling techniques and we answer RQ2 by summarizing our findings with respect to which models are more (or less) sensitive to the choice of the scaling technique. 

Concerning RQ3, in Section \ref{sec:rq3} we look for a relation between the behavior of a monolithic model and an homogeneous ensemble built with it, regarding how the scaling techniques rank for that specific model. Finally, in Section \ref{sec:mean_range_vs_avg_ranking} we present our findings on how the sensitivity to the choice of scaling technique relates to the model's performance, and we discuss how this analysis can be useful for model selection in different deployment scenarios.

\subsection{Models performances}
\label{sec:answ_rq1_rq2}

In order to answer the \textbf{research question RQ1} -- Does the choice of scaling technique matters for classification performance? -- for each dataset, we compared the classification performances of each model when they were trained with the five scaled copies of the dataset and the baseline, which is the original, nonscaled dataset. We then applied a Friedman hypothesis test to evaluate the statistical significance of the differences presented.

Table \ref{tab:mean_performances} show, for each one of the 20 classification models, the values obtained for F1 and G-Mean for each scaling technique after taking the mean across all datasets. In these tables, we abbreviated the scaling techniques names as NS - No scaling (baseline), SS - Standard Scaler,  MM - Min-Max Scaler, MA - Max Absolute Scaler, RS - Robust Scaler, QT - Quantile Transformer. Full results for the experiments can be found in CSV files in the GitHub repository mentioned above.

\input{tabs/tab_mean_performances}

It is interesting to notice that Table \ref{tab:mean_performances} shows that nonscaled data are not always the least performant method. This finding endorses the need to properly and wisely select the scaling technique for a specific pair of dataset and model, as a carelessly selected technique can be worse than not scaling the data at all. 

Table~\ref{tab:mean_performances} also shows that the performance variation when we consider different scaling techniques seem to be relevant for most (14 out of 20 models), except for three monolithic models: LDA, QDA and DT, for the three ensemble models based on DTs: XGBoost, RF, and AdaBoost.

We performed Friedman tests in order to estimate the statistical significance of the observed differences in performance when using different scaling techniques. The test was performed considering these hypotheses and a 0.05 significance level:

\begin{quote}
    \textbf{$H_0$} - There is no significant difference between the means obtained for models built with datasets processed with  different scaling techniques.
\end{quote}

\begin{quote}
    \textbf{$H_1$} - There is a significant difference between the means obtained for models built with datasets processed with different scaling techniques.
\end{quote}

For each of these tests, each sample being compared is composed of the 82 performance measurements for a certain pair of models and scaling technique, whose means were previously presented in Table \ref{tab:mean_performances}. In order to enable a discussion on the role of class balancing in the observed performance differences, we also performed three other sets of Friedman tests considering datasets in different IR strata: low IR datasets (IR $\le$ 3.0) i.e. virtually balanced datasets, medium IR (3 $<$ IR $<=$ 9) and high IR (IR $>$ 9). Results for all four sets of tests are presented in Table \ref{tab:friedman_diff_ST_low_med_high_all}.

\input{tabs/tab_friedman_diff_ST_low_med_high_all}

From this table, we notice that the tests for balanced datasets (i.e., low IR) and the medium IR datasets show much fewer null hypothesis rejections (15 and 11 rejections out of 40, respectively) than those performed considering all the datasets (28 rejections out of 40) or even just the high IR datasets (27 rejections out of 40). This observation is an indication that, although the scale sensitivity problem exists even for more balanced datasets, highly imbalanced datasets are more prone to significant performance variation due to the choice of different scaling techniques. Additionally, even for balanced datasets, the very low p-values observed for the SVMs, QDA, Percep, MLP and Bagging indicate that those differences in performance should not be neglected.

We now shift our focus to the last two columns in Table \ref{tab:friedman_diff_ST_low_med_high_all}, which correspond to the tests results considering all the 82 datasets. When we take into account the monolithic models (from SVM\_lin to MLP), the results state that the null hypothesis can be rejected in most cases, except for the Decision Tree (DT) model and partially (one of two metrics) for the KNN model. Although in the case of KNN, the 0.07 p-value was close to the 0.05 threshold. This means that, for almost all monolithic models studied, there is a significant difference in performance when different scaling techniques are applied to imbalanced datasets. The insensitivity to scale observed in the DT model was already expected since this is a rule-based model, and it assigns decision thresholds for each attribute independently of the scale of the other attributes.

Although we expected that the KNN model presented more significant performance differences among the different scaling techniques, it is not as affected as several other classifiers such as SVM\_lin, SVM\_RBF MLP, Perceptron and GP. One must remember that this model is only sensitive to scale when the difference among the features significantly changes an instance's set of neighbors. Thus, affecting the final classification. This may not necessarily be the case with most of the studied datasets. 

For the ensemble models, the results show that ensembles built with Decision Trees as their base models (RF, XGBoost e AdaBoost) are also insensitive to the choice of scaling technique. On the other hand, the remaining ensembles, built with Perceptrons as their base models, are sensitive to the scale of the data, with the KNORAE model being a notable exception. This indicates that there seems to be a relation between the sensitivity to scale that we observe in a monolithic model and that of an ensemble built based on that model. This topic is further explored in Section \ref{sec:rq3}, where we answer RQ3.

As for the stability observed in the results from the KNORAE model, a possible explanation is that the algorithm works by selecting all base classifiers with a perfect performance in the region of competence (RoC) of the query instance (its $k$ nearest neighbors), but while no classifiers are attaining perfect performance in the RoC, $k$ keeps being decremented by 1, reducing the region of interest. This way, when the most distant neighbor(s) is(are) eliminated, the algorithm removes samples with extreme values for some features. This way, it reaches a RoC where samples present more well-behaved scales among its features, over which a different choice of scaling technique has little effect.

\begin{figure*}[!ht]
    \centering
    \includegraphics[trim=0 5 0 0,clip,  width=0.97\linewidth]{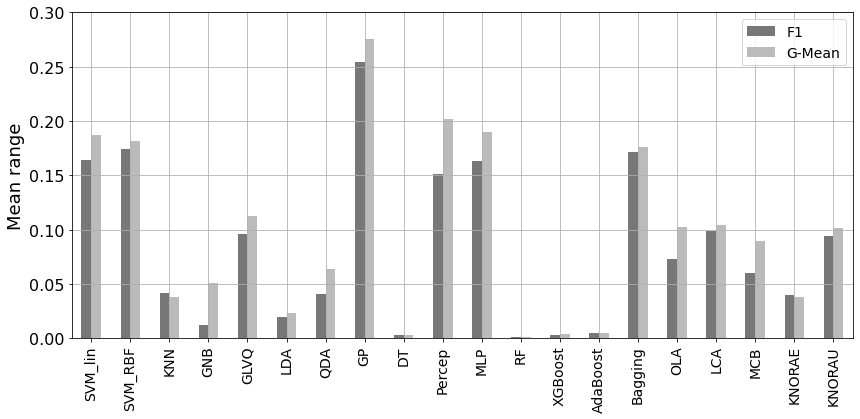}
    \includegraphics[trim=-4 0 4 5,clip,  width=0.97
    \linewidth]{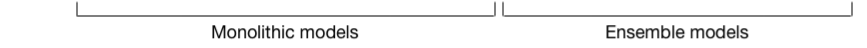}
    \caption{Mean ranges (differences from best to worst scaling technique) for the monolithic and ensemble models.
    }
    \label{fig:barplots_range}
\end{figure*}

In order to better visualize how much difference in performance the choice of scaling technique can promote, in Figure \ref{fig:barplots_range} we show the bar plots representing the performance variation range for each model. The mean over all datasets is considered here. These graphs, as expected, endorses the results and conclusions obtained from the hypothesis test, as we can see greater ranges for the SVMs, GP, Percep and MLP monolithic models and very low ranges for the DT model and the ensembles built with it as a base model.

Additionally, notice that, for the monolithic models, the mean difference in performance (mean range) is close to, or more than, 0.15 for five models (SVM\_lin, SVM\_RBF, GP, Percep and MLP), and this can represent, in some contexts, the difference between a useful and a useless model. For the ensemble models, the differences are less impressive but not negligible.

Provided all the above evidence, we can answer the research question \textbf{RQ1: Yes, the choice of scaling technique matters for classification performance.} We could verify this for both monolithic and ensemble models. The hypothesis tests show that the performance differences are significant for most models. The varying extent to which scaling affects performance for distinct models brings us to our next research question.


With the same resources we employed to answer RQ1, we can also answer \textbf{research question RQ2} -- Which models present greater performance variations when datasets are scaled with different techniques? -- we begin looking back at Figure \ref{fig:barplots_range}, which presents the mean ranges (considering all 82 datasets) in model performance (difference between the best and worst scaling technique). As previously discussed, there appears to be a great variability for some models in both monolithic and ensemble groups and almost no variation when considering a few select models.

It is important to stress that Figure \ref{fig:barplots_range} presents the means over all 82 datasets. Then, for the most sensitive models, the performance gain when choosing the right scaling technique may be even higher for some datasets. If we take the SVM\_lin model, for example, looking at the raw results (available in the supplemental materials), we can see that its maximum difference between the best and the worst scaling technique occurred for the ``segment-0'' dataset, a range of 0.991 for the F1 metric. This result for the ``segment-0'' dataset is not exactly an outlier, since, for the SVM\_lin model, the range achieved in 26 (~32\%) of the 82 datasets was above 0.5, and in 6 (~7\%) of them, the range was higher than 0.8. These are extreme differences between the best and worst performances of the same model caused only by swapping the scaling techniques. This further emphasizes the need for careful choice of the scaling technique to be applied in a classification task.

We can see in Figure \ref{fig:barplots_range} that some models, such as SVM\_lin, SVM\_RBF, GLVQ, GP, Percep, MLP, Bagging, OLA, LCA and KNORAU present larger mean ranges while models such as DT, and the DT-based ensembles (RF, XGBoost and AdaBoost) are almost insensitive to the choice of scaling technique. This is confirmed by the results of the Friedman tests in Table \ref{tab:friedman_diff_ST_low_med_high_all}, which leads us to some possible generalizations: Considering the monolithic models and their categories in Table \ref{tab:params}, instance-based (KNN is an exception) and Neural Networks (MLP and Percep) models seem to be more sensitive to how data are scaled. In contrast, rule-based and discriminant analysis models are less sensitive. When we consider the ensemble models, those built using decision trees as their base models (XGBoost, RF and AdaBoost) are equally insensitive to the scaling technique chosen, while those based on Perceptrons (all the others) are more sensitive.

As an additional analysis, in Table \ref{tab:st_wins_per_ir_stratum} we present the number of wins of each scaling techniques (how many times each one was better than the others) for each IR strata. This allows us to conclude that considering all the IR strata the Quantile Transformer (QT) technique is the most successful. In the low IR stratum, the best technique is the Standard Scaler (SS), while in the medium and high strata the Quantile Transformer is again the most performant technique. This indicates that the IR may be an important variable to determine which scaling technique is the best for a certain dataset.

\input{tabs/tab_st_wins_per_ir_stratum}

\subsection{Do scaling techniques rank similarly for an ensemble and its base model?}
\label{sec:rq3}

Up to this point, we can see from Figure \ref{fig:barplots_range} that there is an apparent relationship between the variability to scale observed in ensembles and that of their base models. This observation demands further investigation that we intend to carry out as we answer \textbf{research question RQ3} - Do homogeneous ensembles tend to follow the performance variation pattern presented by their base model when dealing with different scaling techniques?

By ``performance variation pattern'' we mean the order in which the scaling techniques rank when considering a particular model and dataset. We then aim to answer if, on average, the best ranking scaling technique is the same between a specific monolithic model and its corresponding ensemble and also if the second, third-ranking techniques and so on, are coincident.

In order to better compare these rankings for a monolithic model and an ensemble built with it, we must select a monolithic model and all its corresponding ensembles. Due to the way we designed the ensembles in our study, we have two possible monolithic models for this study: DT or Percep, because the ensembles we built have either DT or Percep as their base models. Since we have already verified that the DT model and its correspondent ensembles are practically scale insensitive, we chose to compare the Percep monolithic model and the ensembles built with it as their base model: Bagging, OLA, LCA, MCB, KNORAE, KNORAU. 

Although we are limiting this analysis to perceptrons and perceptron-based ensembles, this is a specially useful model for building ensembles due to its simplicity, low computational cost and sensitivity to data sampling, which helps promote the much-desired ensemble diversity \cite{Kuncheva2014}. Additionally, it has been shown in the DES literature that the use of weak models as a base classifier, such as decision trees and perceptrons, allows better results, outperforming stronger models like Random Forest, SVM or MLP \cite{Cavalin2010, Cavalin2013, Souza2019, Cruz2015MetaDES, Cruz2015METADES.H}.

\begin{figure*}[!ht]
    \centering
    \includegraphics[trim=0 75 0 0,clip,  width=0.25\linewidth]{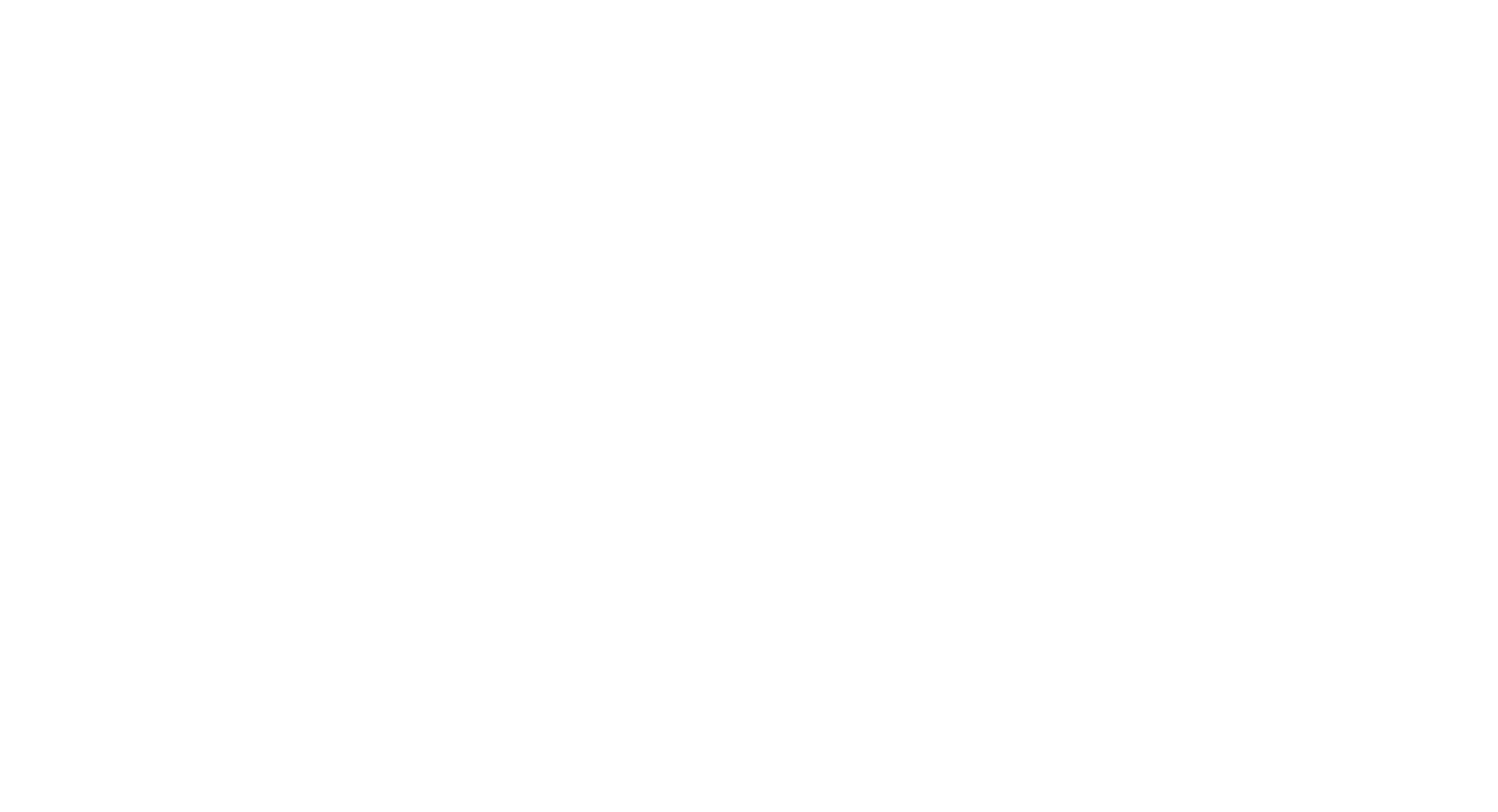}
    \includegraphics[trim=0 75 0 0,clip,  width=0.45\linewidth]{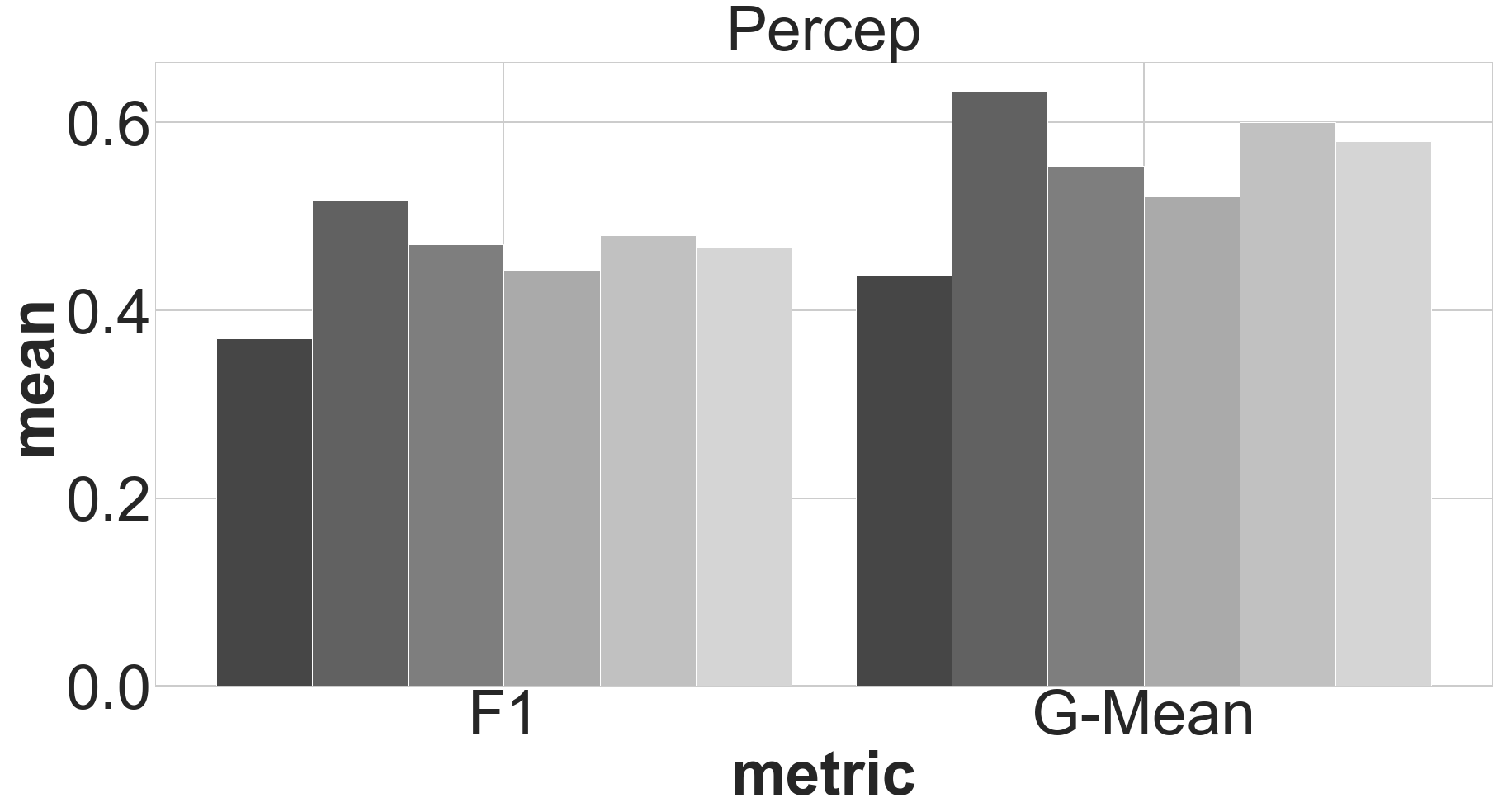}
    \includegraphics[trim=0 75 0 0,clip,  width=0.25\linewidth]{figs/white_fill.png}
    
    \includegraphics[trim=0 75 0 -40,clip,  width=0.45\linewidth]{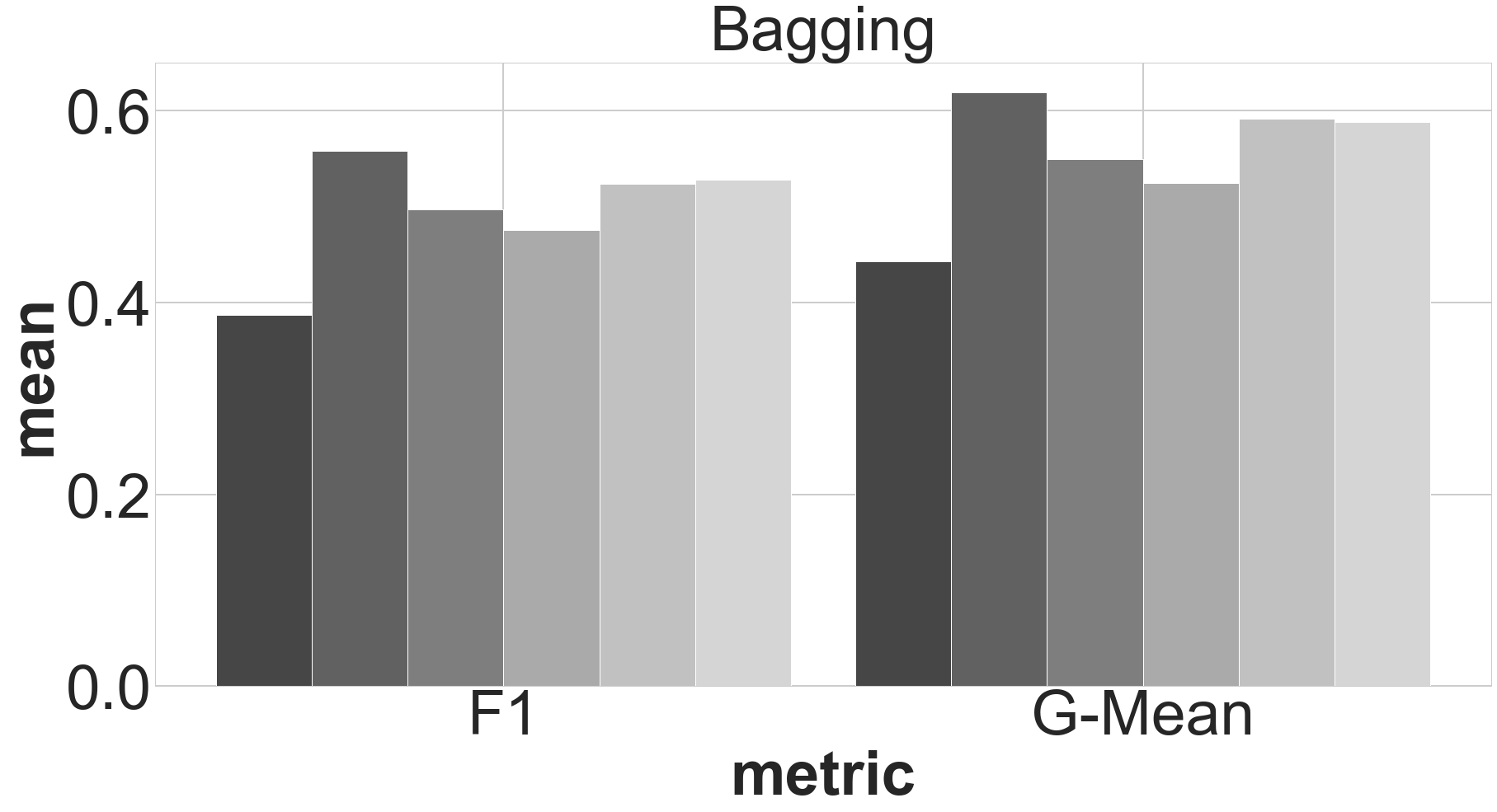}
    \includegraphics[trim=0 75 0 -40,clip,  width=0.45\linewidth]{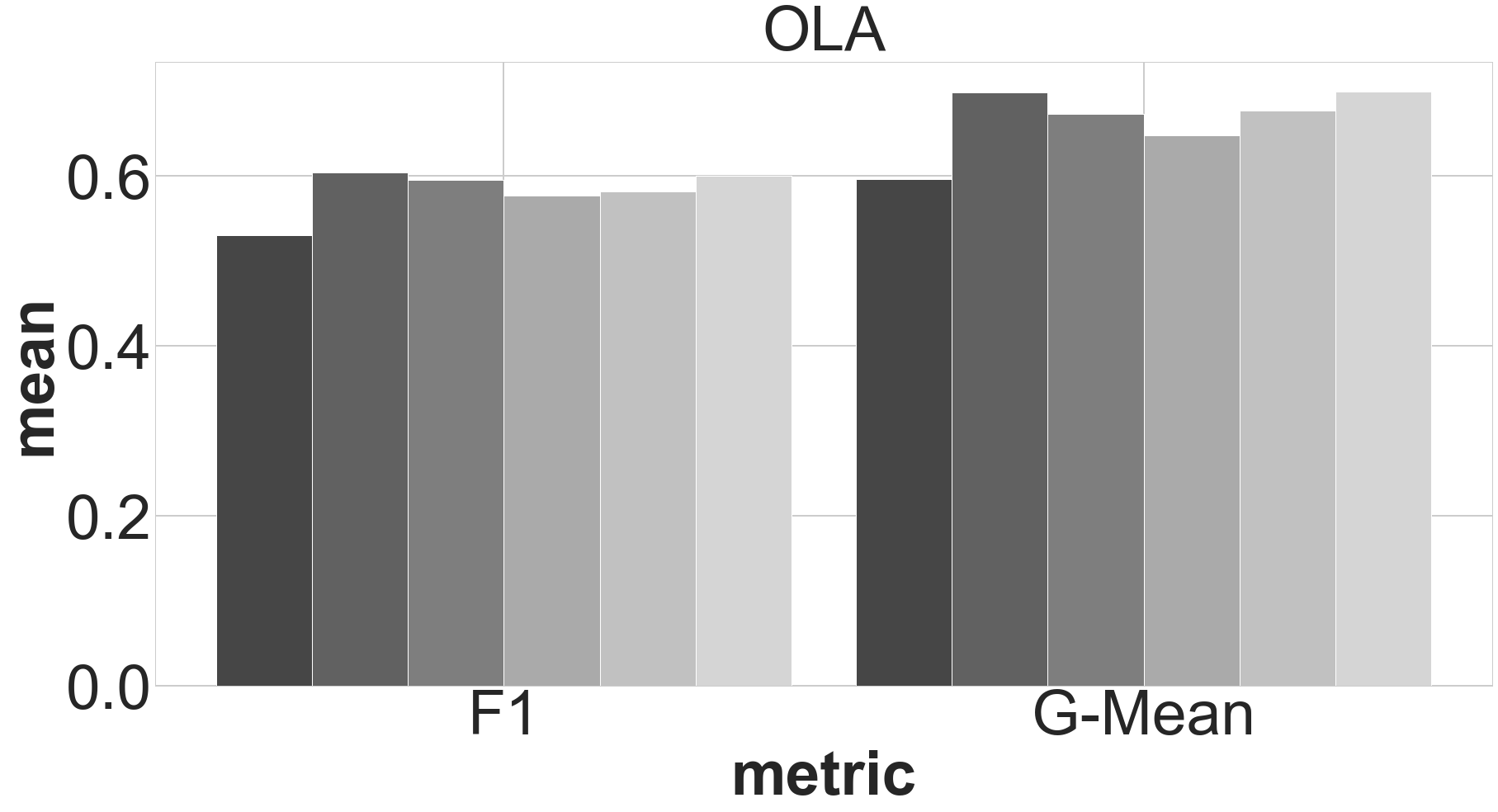}
    
    \includegraphics[trim=0 75 0 -40,clip,  width=0.45\linewidth]{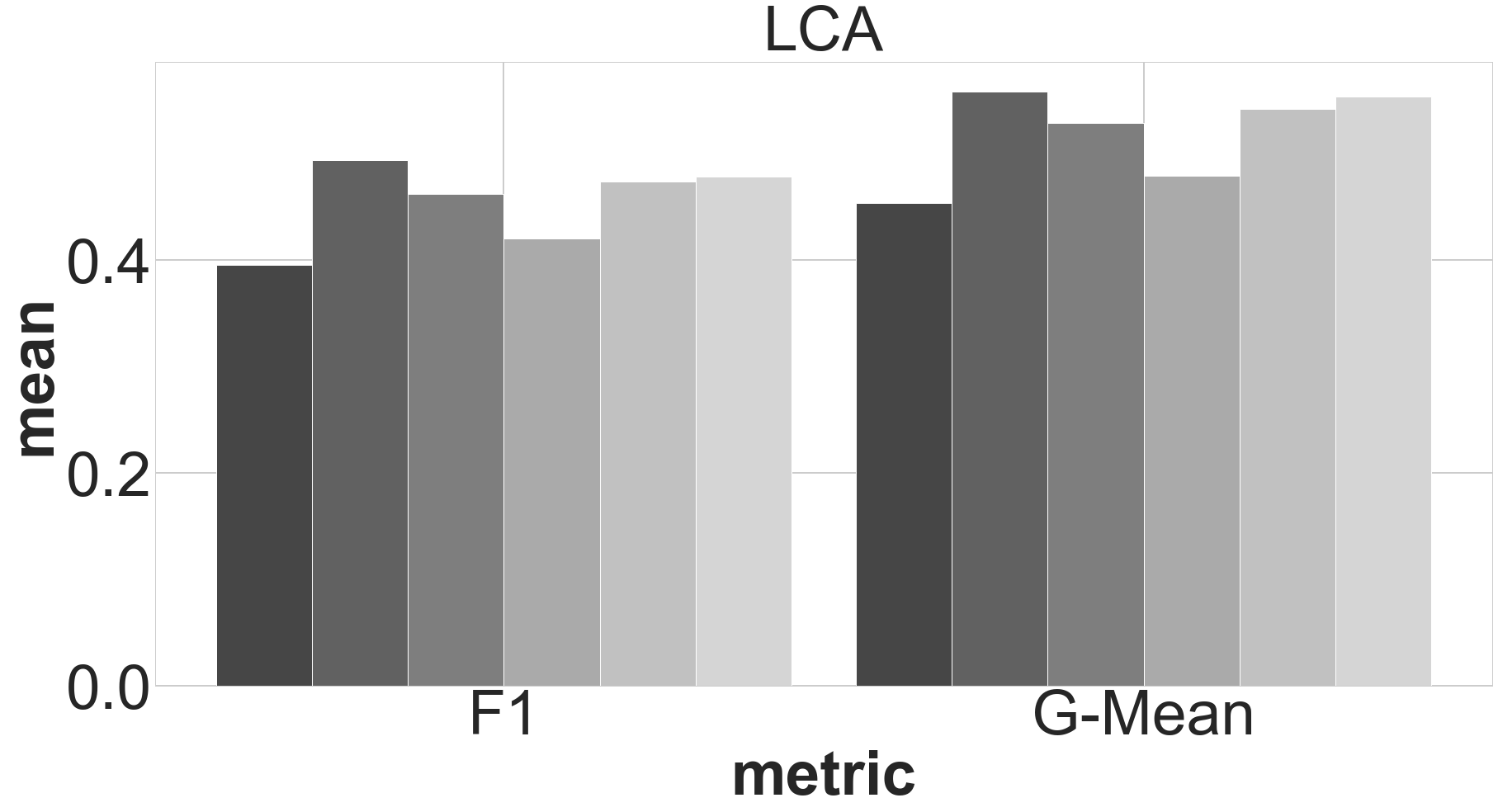}
    \includegraphics[trim=0 75 0 -40,clip,  width=0.45\linewidth]{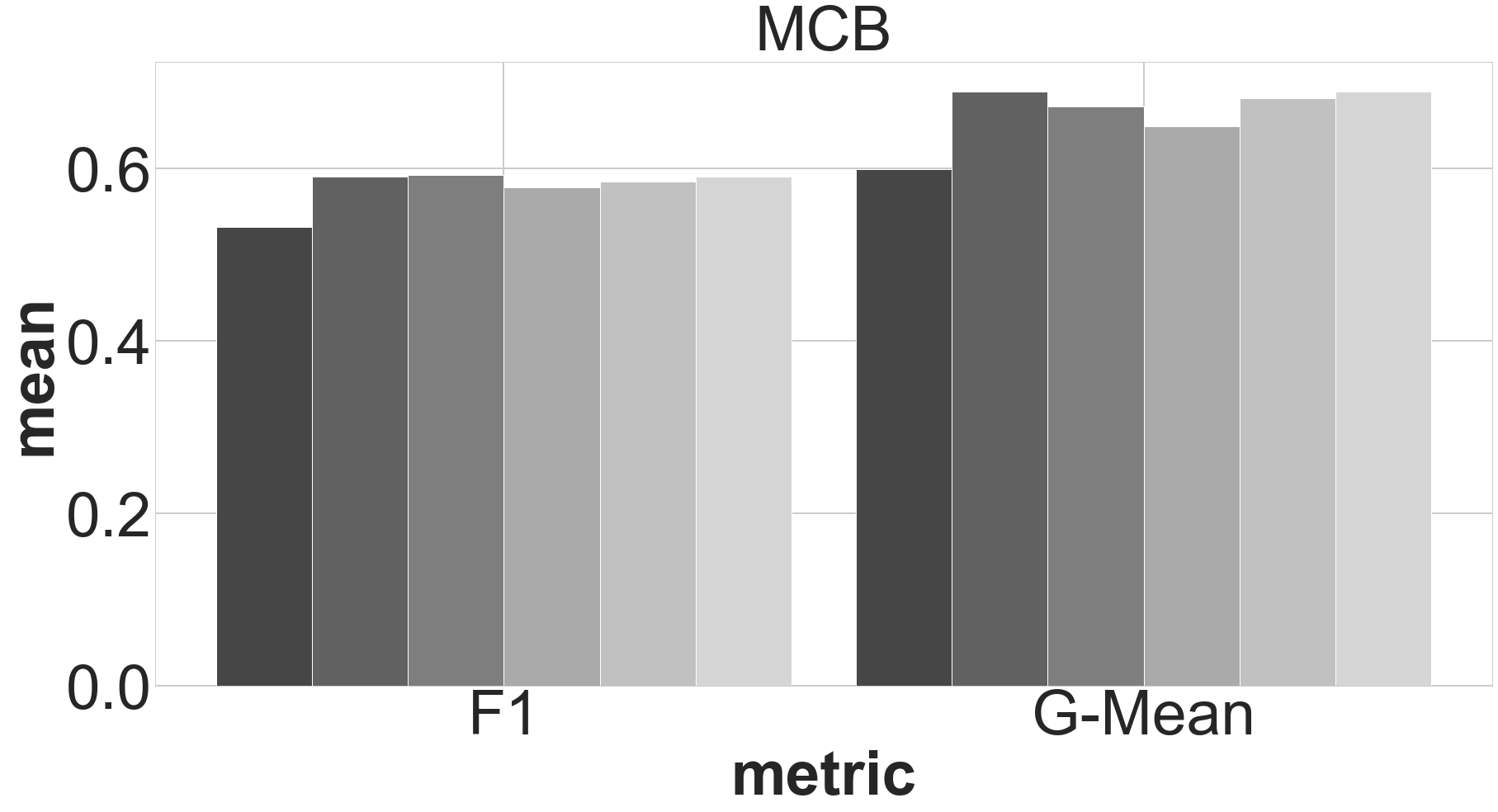}

    \includegraphics[trim=0 75 0 -40,clip,  width=0.45\linewidth]{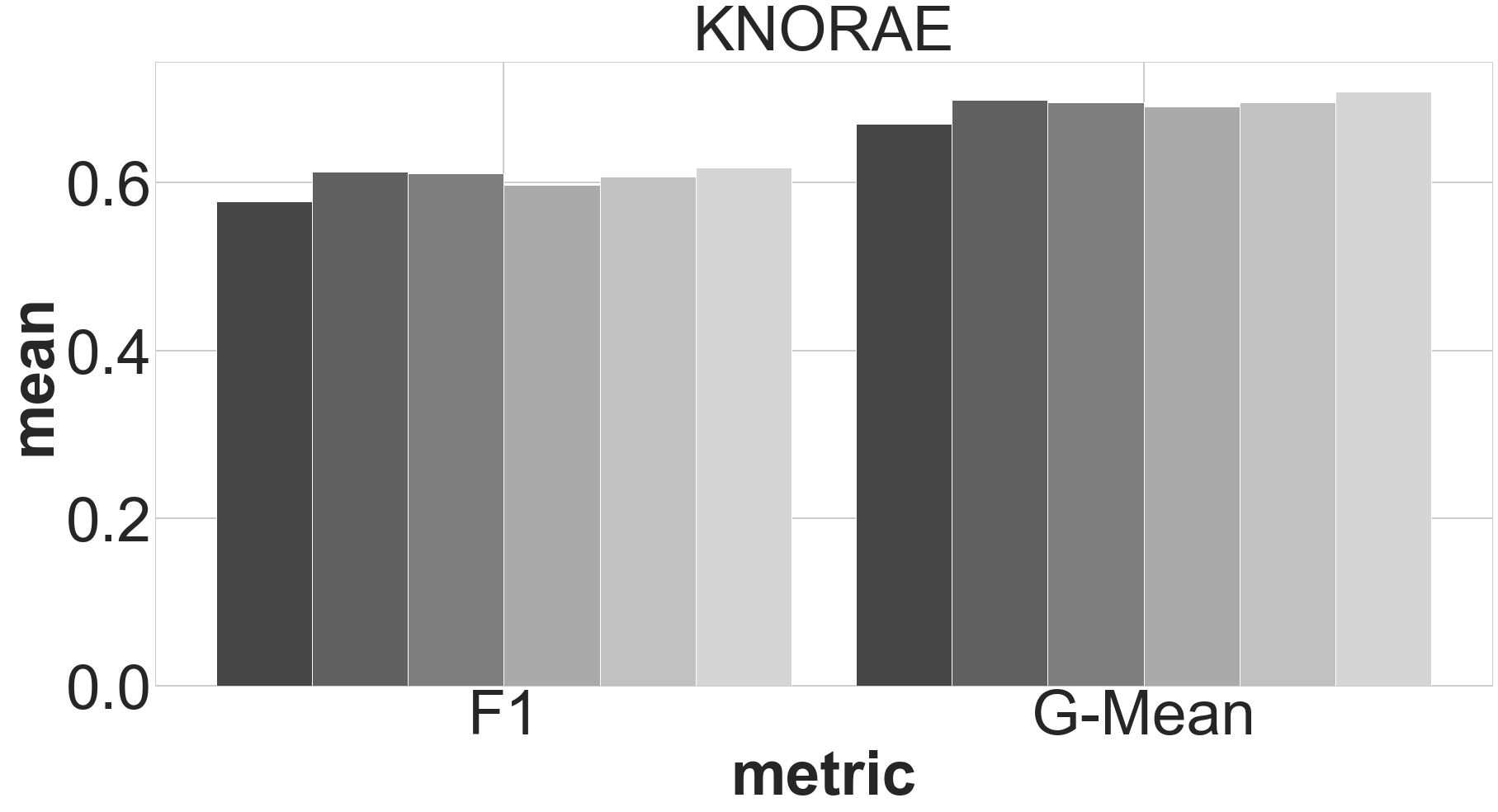}
    \includegraphics[trim=0 75 0 -40,clip,  width=0.45\linewidth]{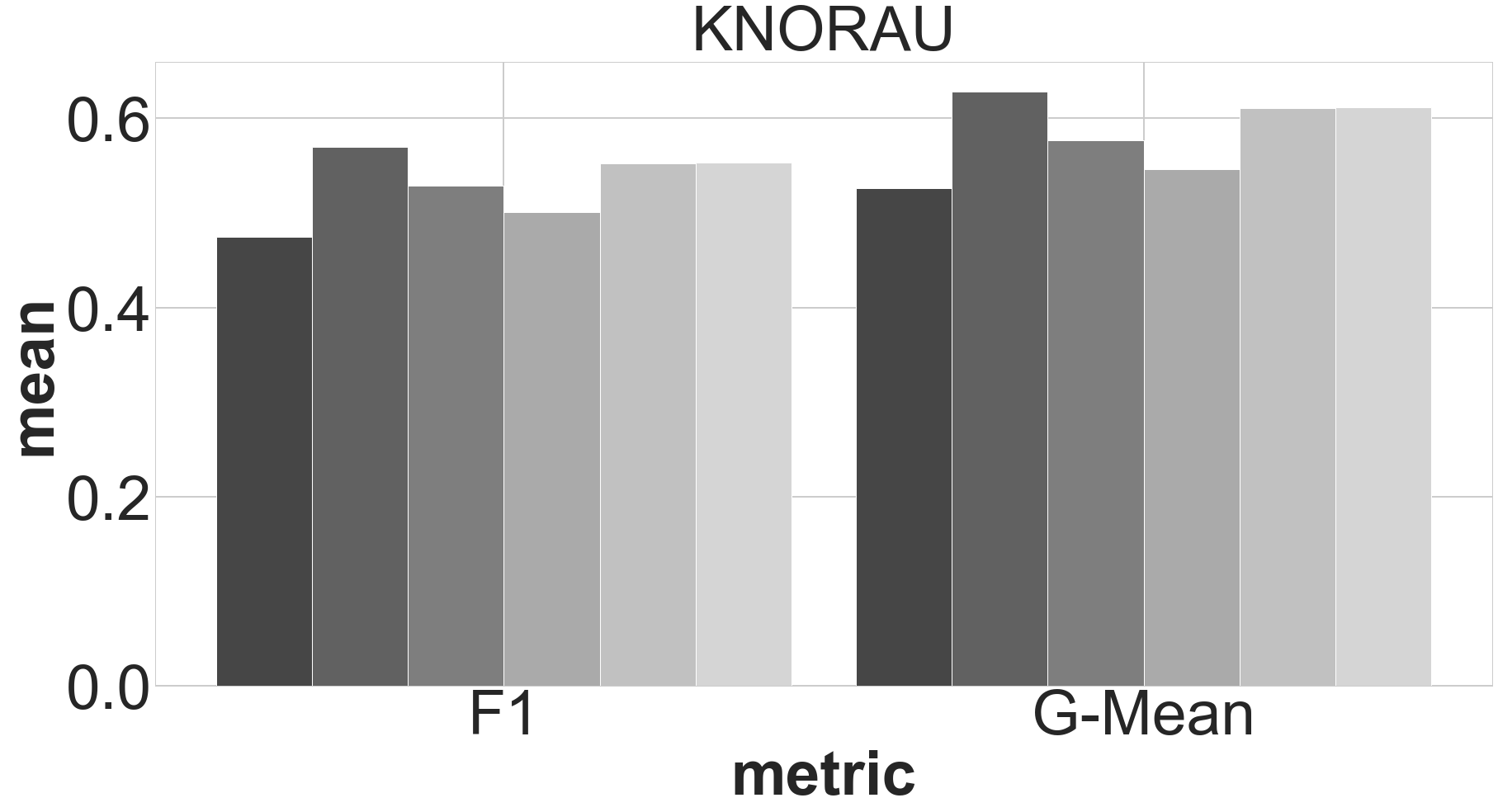}
    
    \includegraphics[trim=0 0 0 -10,clip, width=0.99\textwidth]{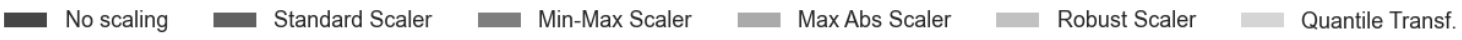}
    
    \caption{Mean performances for the six scaling techniques for the Perceptron model (above) and its correspondent ensembles.}
    \label{fig:barplot_all_ds_by_model_percep_v_ens}
\end{figure*}

Figure \ref{fig:barplot_all_ds_by_model_percep_v_ens} shows, for the Percep and the six ensembles based on it, the performance variation pattern along with the different scaling techniques for both F1 and G-Mean metrics. Each bar represents the mean result over all the 82 datasets for the corresponding scaling technique and metric. 

\begin{figure*}[!ht]
    \centering
    \includegraphics[trim=70 0 70 0,clip, 
    width=0.245\linewidth]{figs/white_fill.png}
    \includegraphics[trim=70 0 70 0,clip,  width=0.47\linewidth]{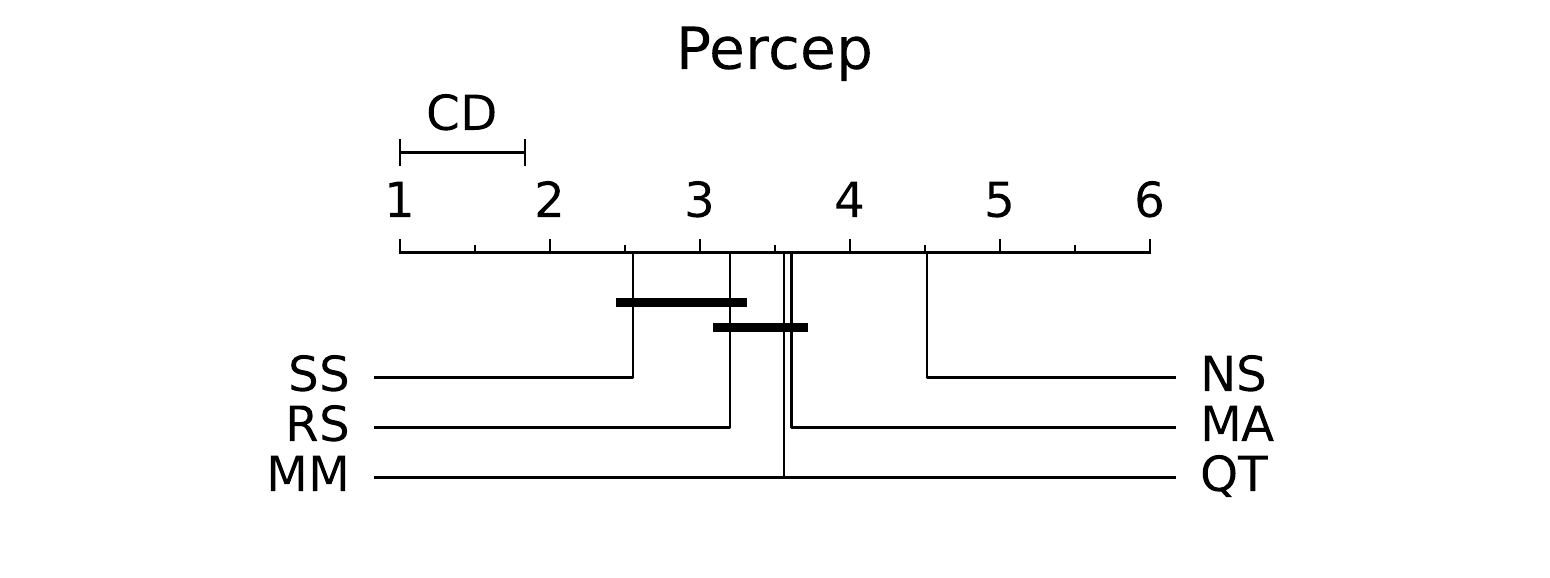}
    \includegraphics[trim=70 0 70 0,clip, 
    width=0.245\linewidth]{figs/white_fill.png}
    
    \includegraphics[trim=70 0 70 0,clip,  width=0.47\linewidth]{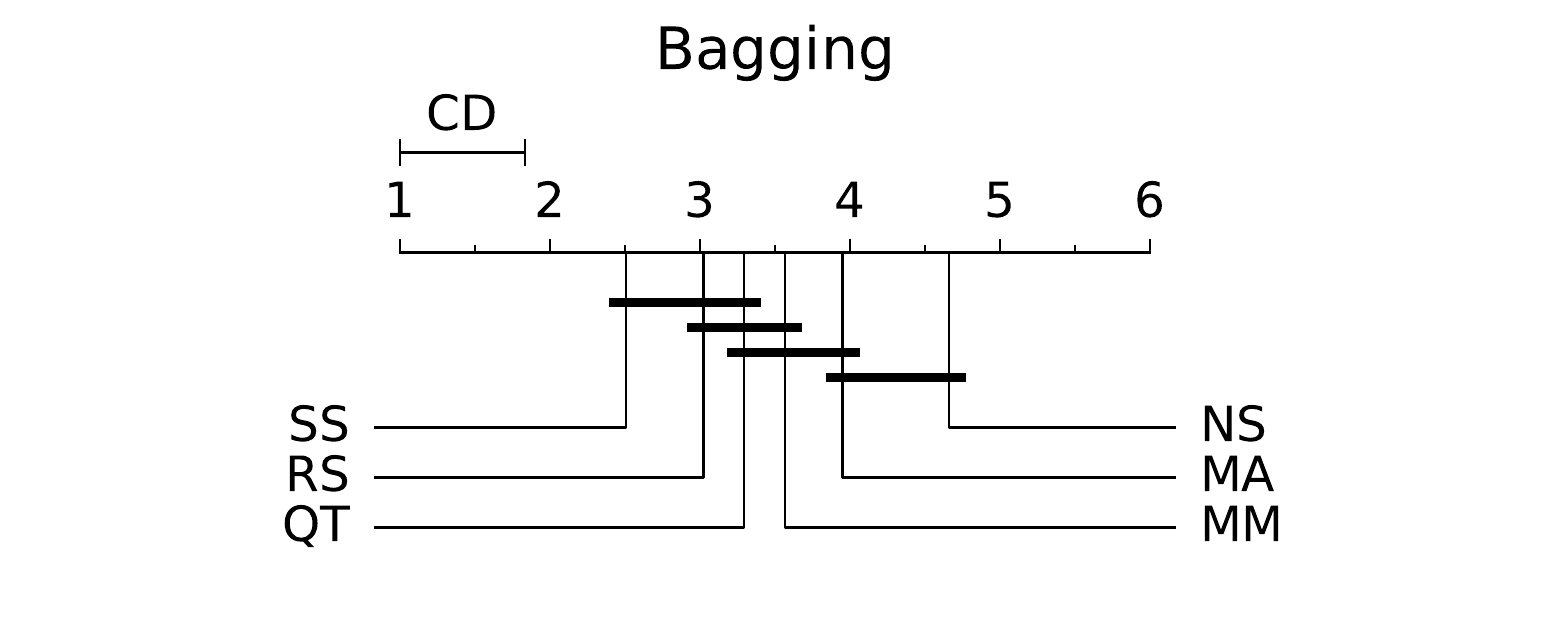}
    \includegraphics[trim=70 0 70 0,clip,  width=0.47\linewidth]{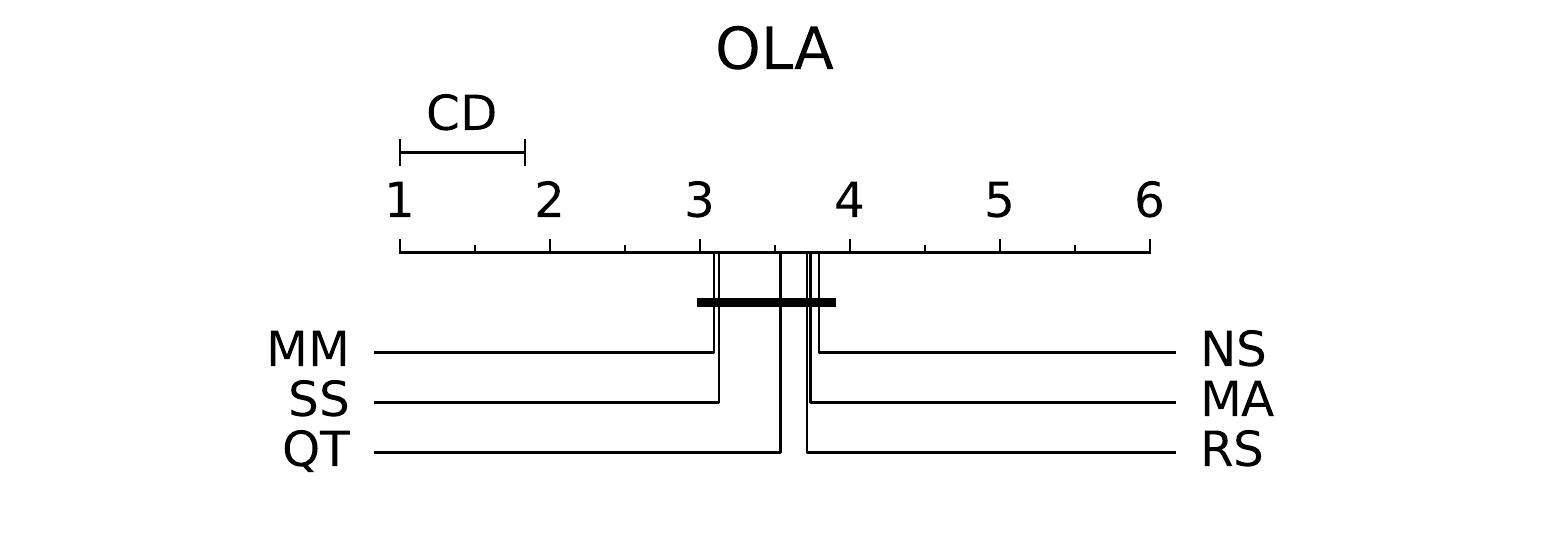}
   
    \includegraphics[trim=70 0 70 0,clip,  width=0.47\linewidth]{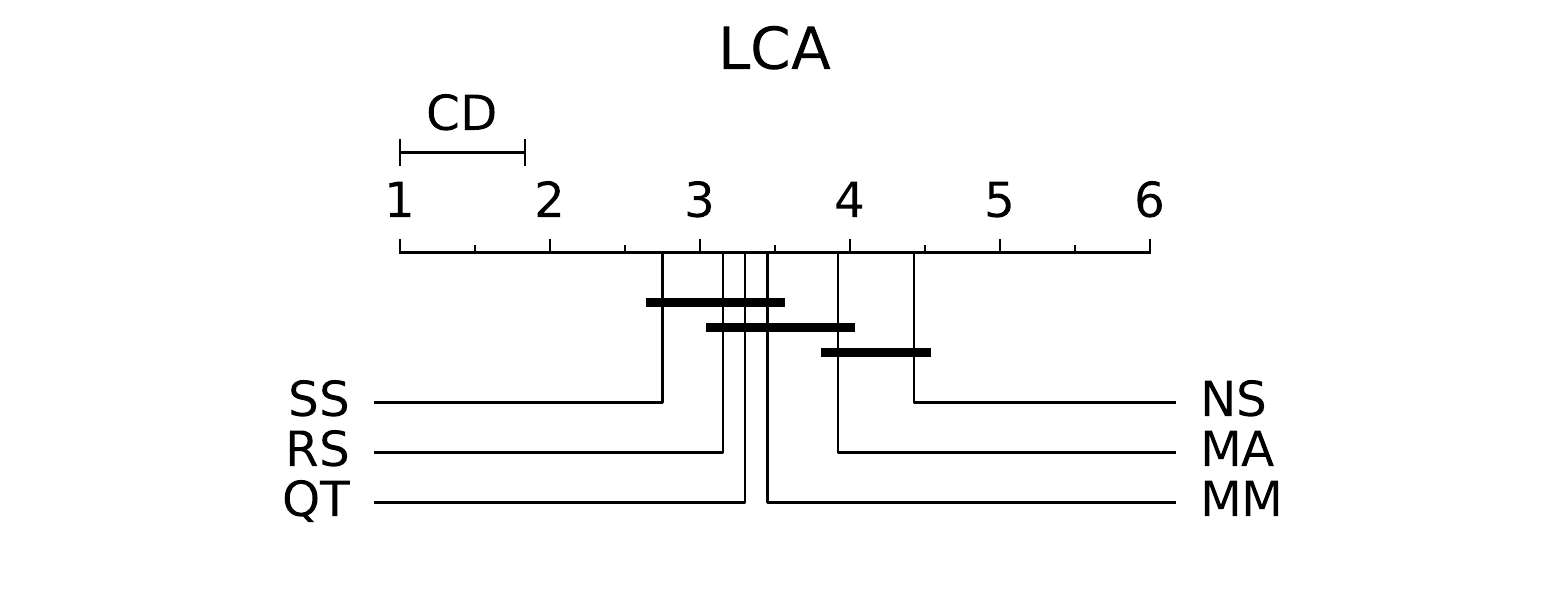}
    \includegraphics[trim=70 0 70 0,clip,  width=0.47\linewidth]{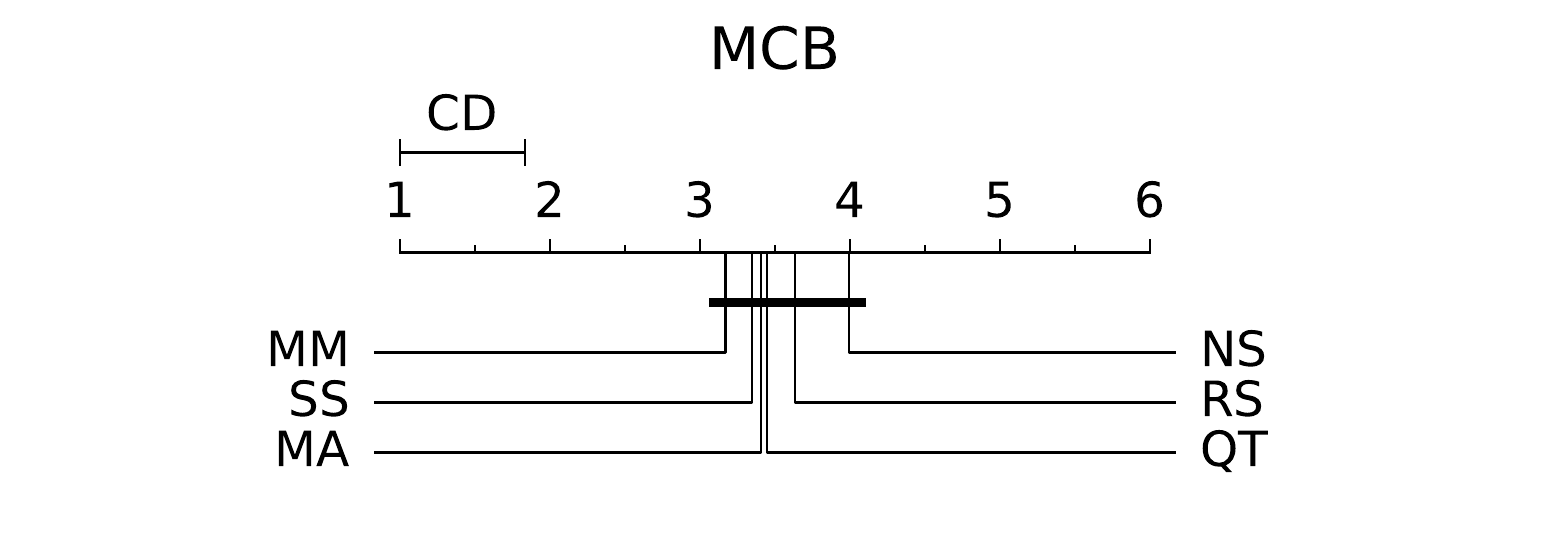}
    
    \includegraphics[trim=70 0 70 0,clip,  width=0.47\linewidth]{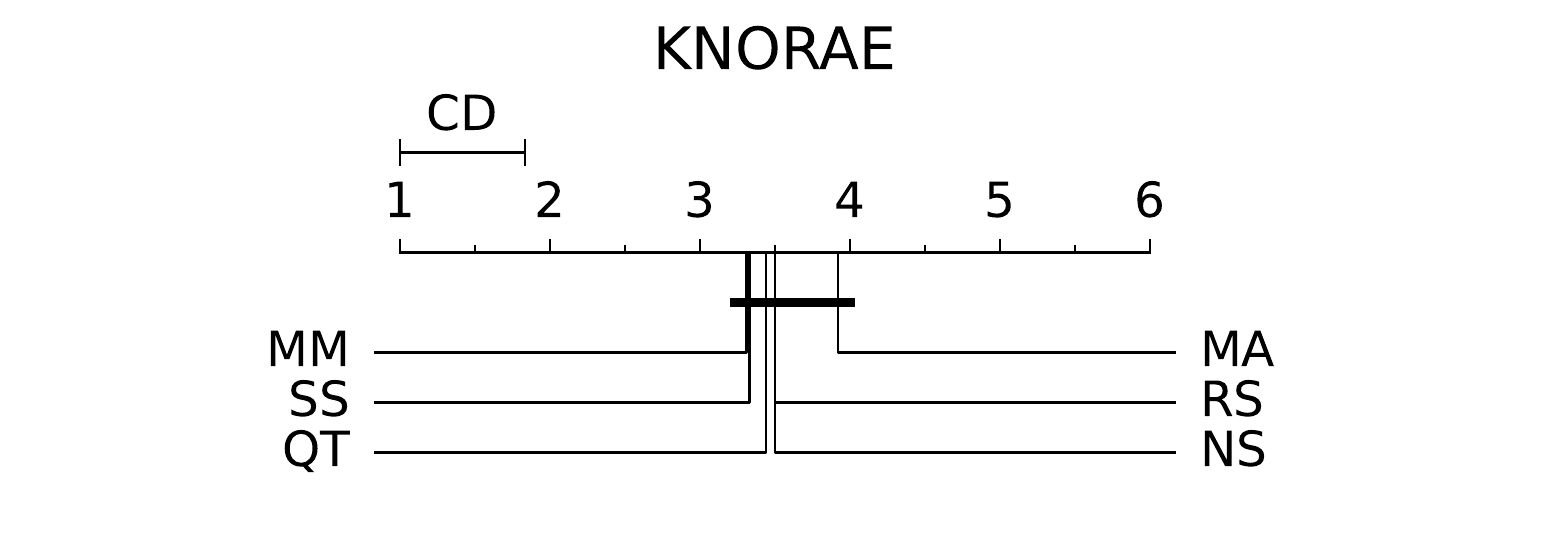}
    \includegraphics[trim=70 0 70 0,clip,  width=0.47\linewidth]{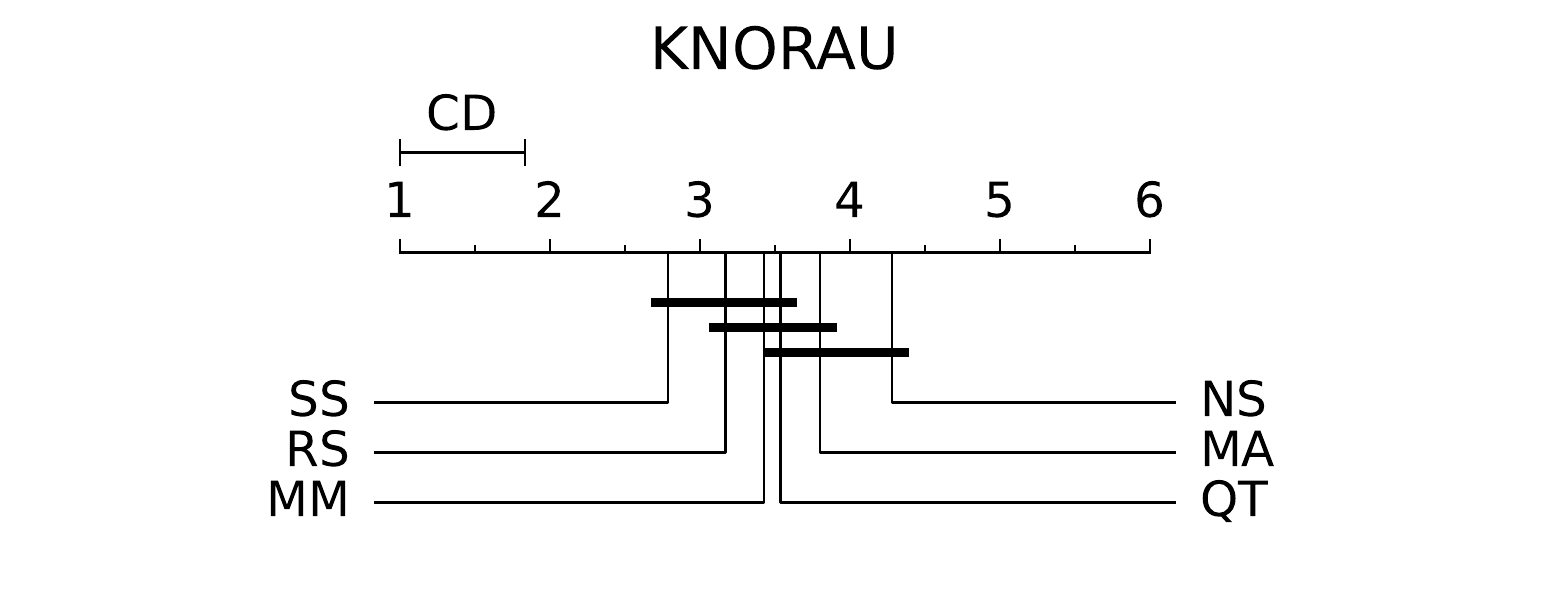}
    
    \caption{CD diagrams of average rankings (considering \textbf{metric F1}) of the six scaling techniques for the Perceptron model (above) and its corresponding ensembles.}
    \label{fig:CD_diagrams_avranks_percep_v_ens_f1}
\end{figure*}

\begin{figure*}[!ht]
    \centering
    \includegraphics[trim=70 0 70 0,clip,  width=0.245\linewidth]{figs/white_fill.png}
    \includegraphics[trim=70 0 70 0,clip,  width=0.47\linewidth]{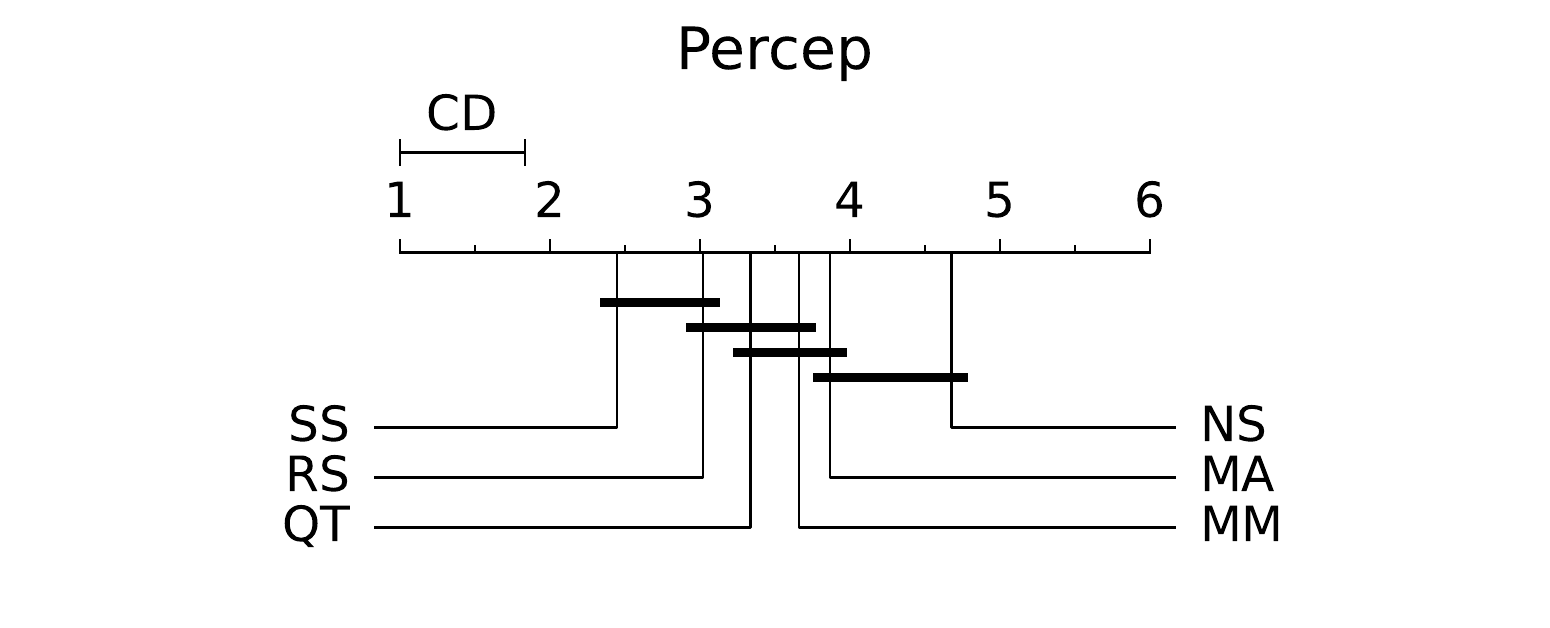}
    \includegraphics[trim=70 0 70 0,clip,
    width=0.245\linewidth]{figs/white_fill.png}
    
    \includegraphics[trim=70 0 70 0,clip,  width=0.47\linewidth]{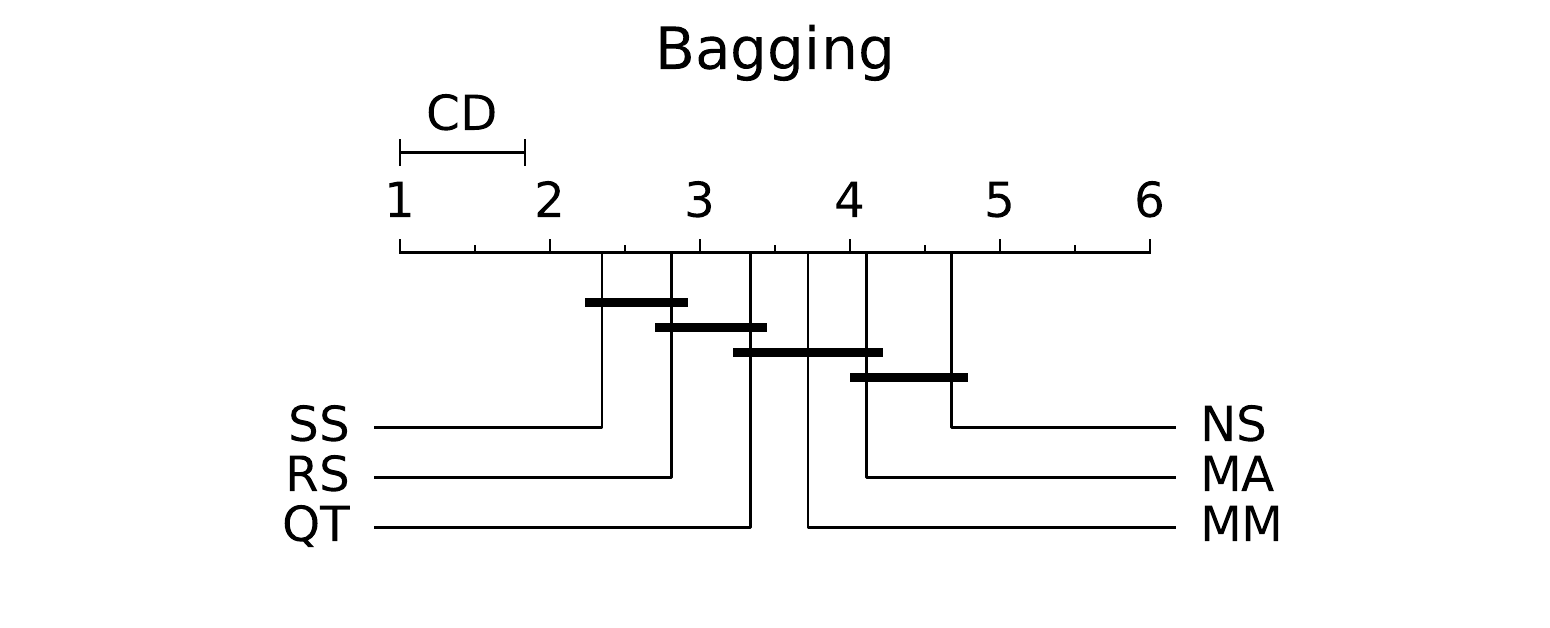}
    \includegraphics[trim=70 0 70 0,clip,  width=0.47\linewidth]{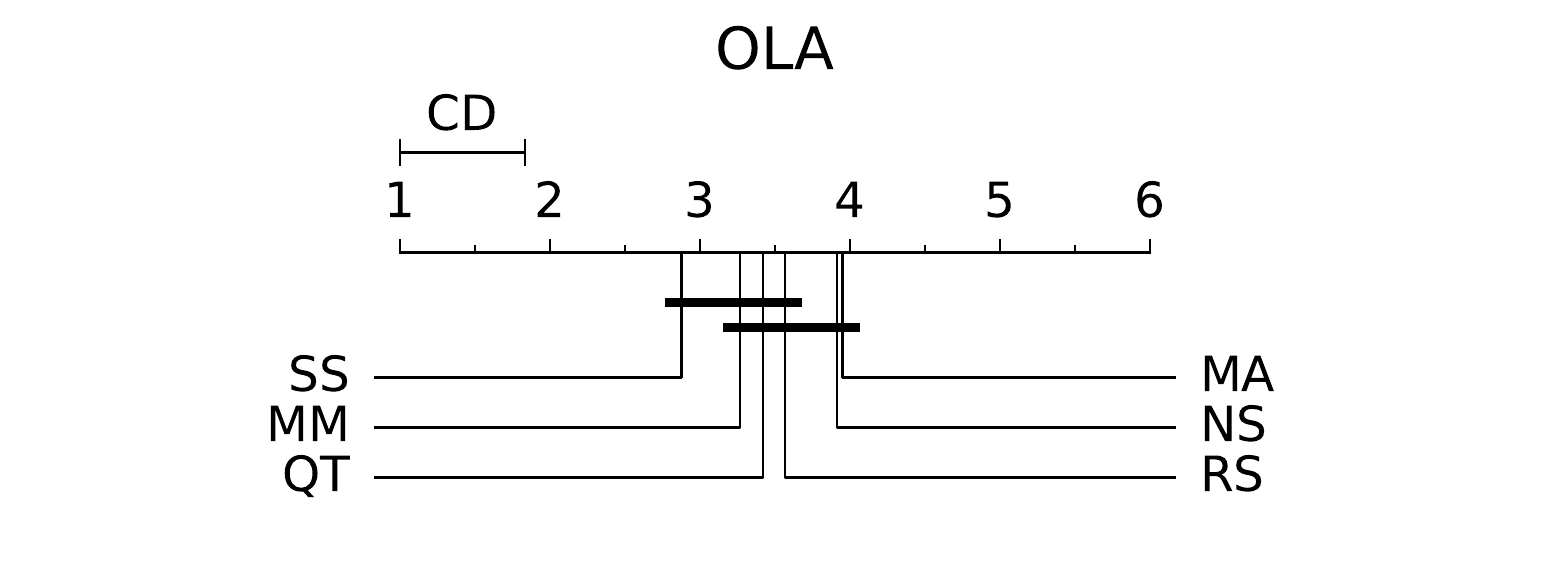}
    
    \includegraphics[trim=70 0 70 0,clip,  width=0.47\linewidth]{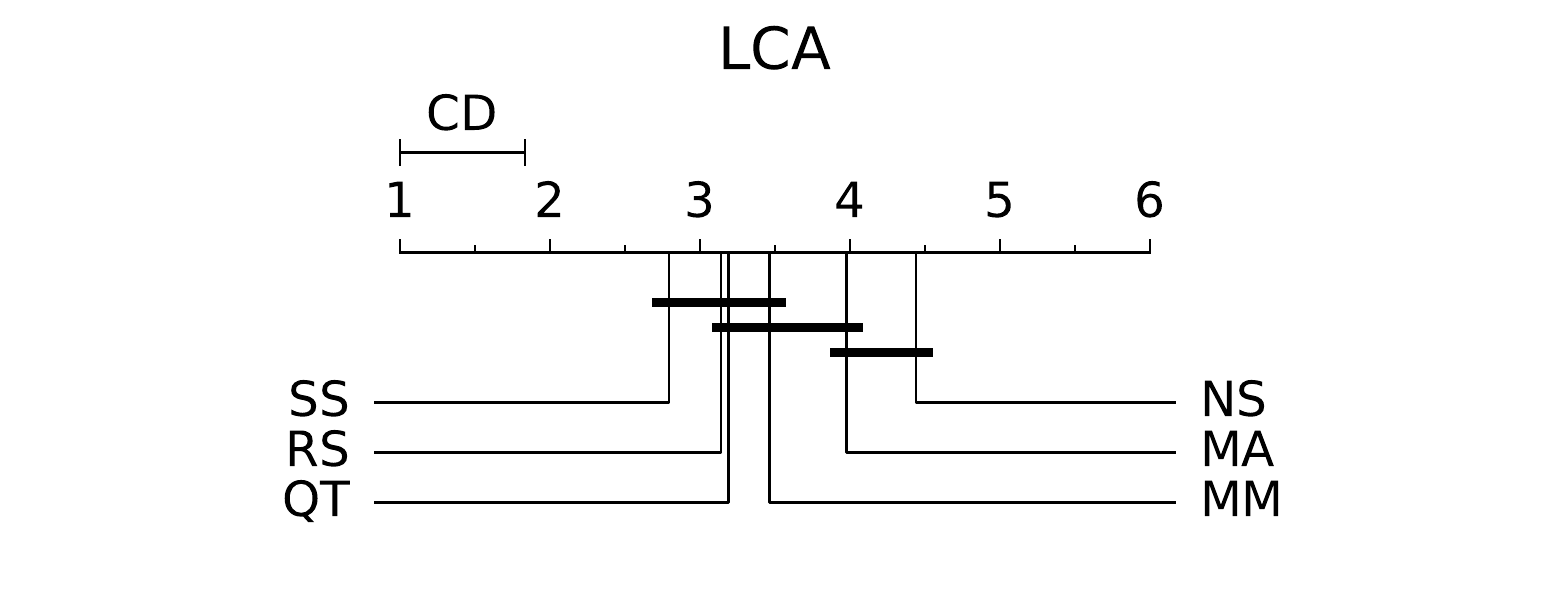}
    \includegraphics[trim=70 0 70 0,clip,  width=0.47\linewidth]{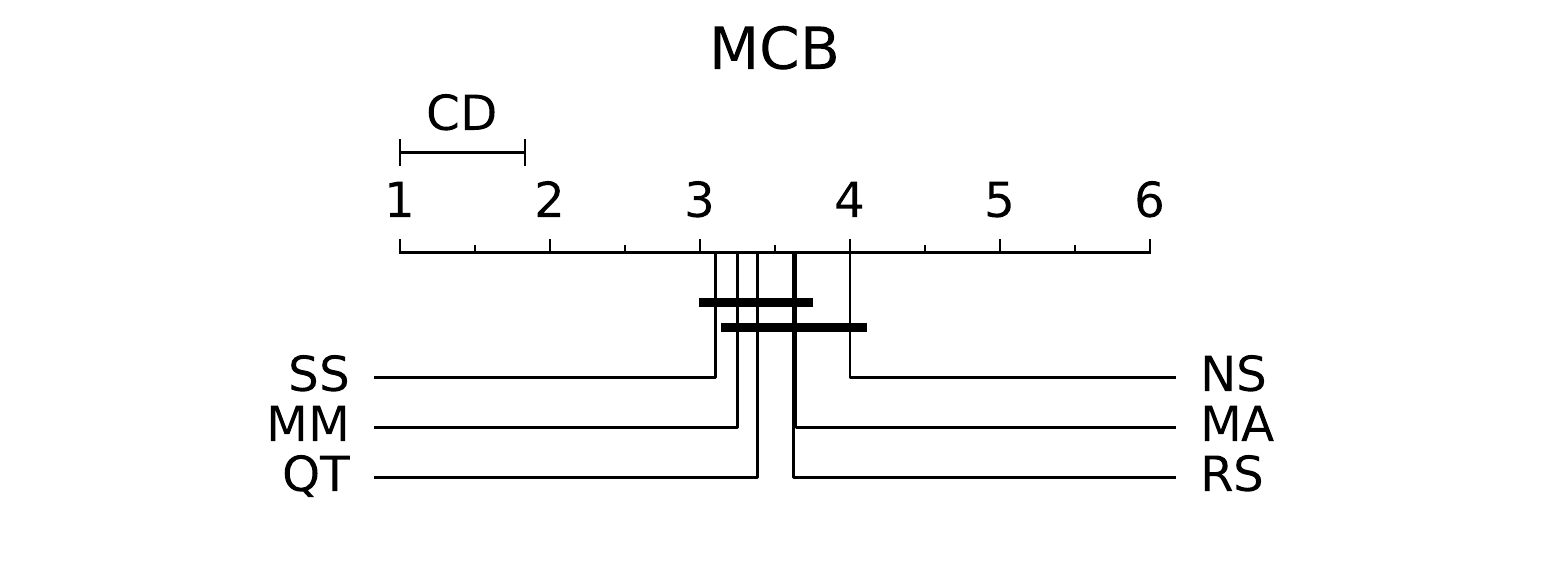}
    
    \includegraphics[trim=70 0 70 0,clip,  width=0.47\linewidth]{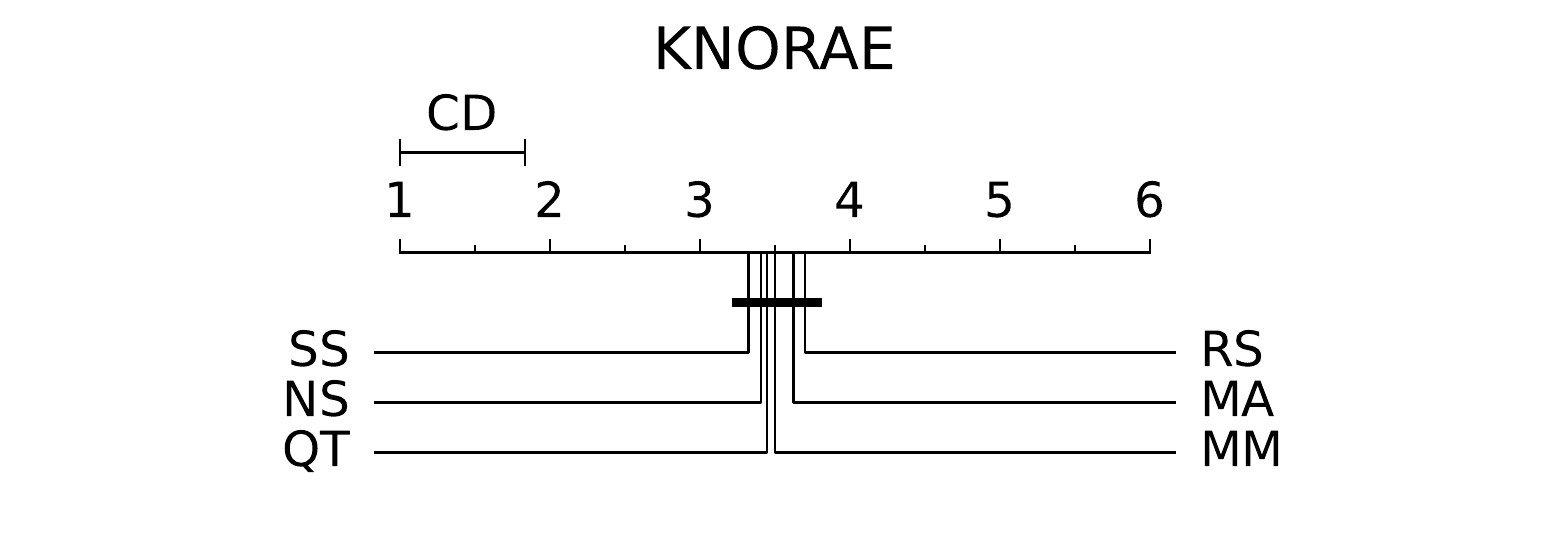}
    \includegraphics[trim=70 0 70 0,clip,  width=0.47\linewidth]{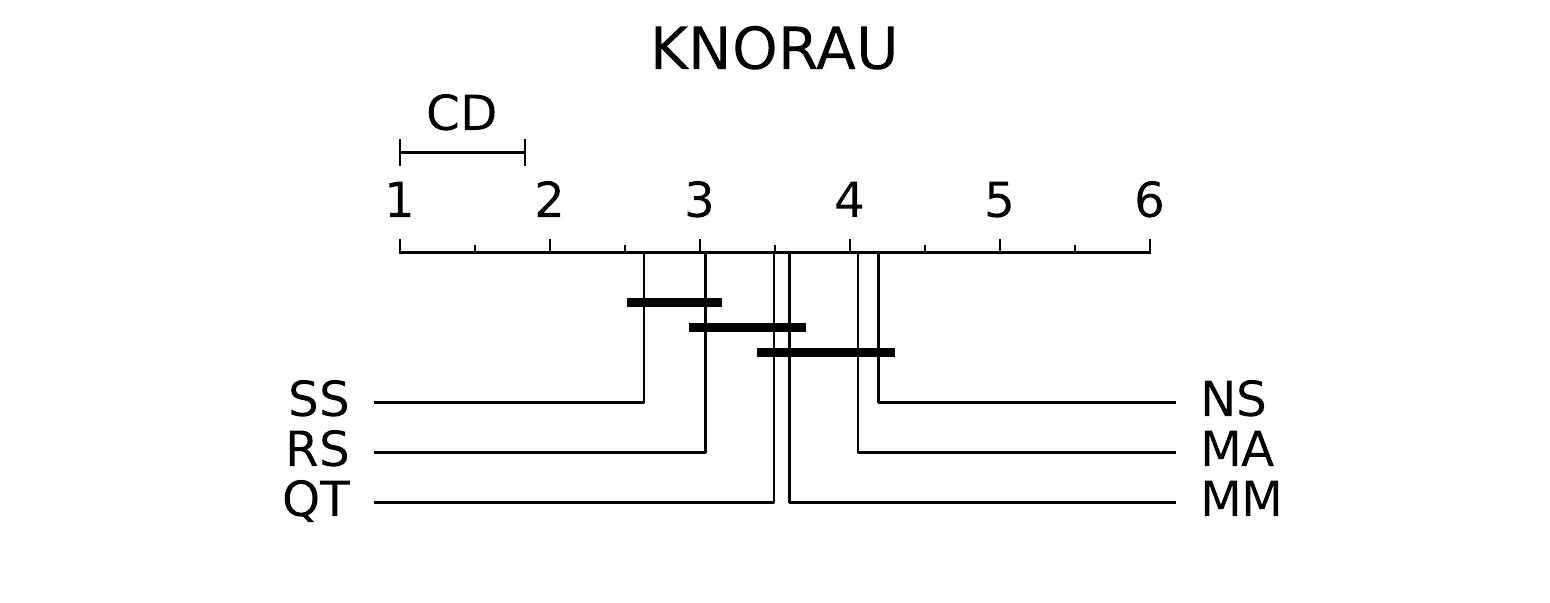}
    
    \caption{CD diagrams, using Nemenyi test, of average rankings (considering \textbf{metric G-Mean}) of the six scaling techniques for the Perceptron model (above) and its corresponding ensembles.}
    \label{fig:CD_diagrams_avranks_percep_v_ens_gmean}
\end{figure*}

We note from these bar plots that, for both metrics, the best and worst scaling techniques are coincident for all models: The best performances were obtained with the Standard Scaler, and the worst results were obtained with nonscaled data (NS). That furthers our belief in the relationship between the behavior of the Percep model and its derived ensembles. However, we can see counterexamples when we look for the bar that ranks second in each model: While that is Robust Scaler for the Percep, it is Quantile Transformer for all the others, although they do not seem to differ significantly. These observations indicate that a better analysis must be made in order to answer this question. Instead of only looking at the rankings of the means in Figure \ref{fig:barplot_all_ds_by_model_percep_v_ens}, we calculated the average ranking (considering the 82 datasets) of each scaling technique for the Percep model versus those from the six corresponding ensemble models and represented the results as critical difference (CD) diagrams in Figures \ref{fig:CD_diagrams_avranks_percep_v_ens_f1} and \ref{fig:CD_diagrams_avranks_percep_v_ens_gmean} for F1 and G-Mean respectively. These CD diagrams were built using a Nemenyi post hoc test.

In Figures \ref{fig:CD_diagrams_avranks_percep_v_ens_f1} and \ref{fig:CD_diagrams_avranks_percep_v_ens_gmean}, we observe that the more scale-sensitive ensembles (i.e., those that are more prone to changes due to the choice of scaling technique) tend to follow the trend set by the Percep model in which they are based. For example, for both F1 and G-Mean, the top two and bottom two scaling techniques are coincident when we consider the Percep and the more sensitive ensemble models (Bagging, LCA and KNORAU): SS and RS appear in first and second places while MA and NS appear as fifth and sixth places.   Nonetheless, this does not happen for models that are less sensitive to scale (specially KNORAE), mainly because, for these models, the differences between the rankings of the distinct scaling techniques are less significant, as can be seen in Figures \ref{fig:CD_diagrams_avranks_percep_v_ens_f1} and \ref{fig:CD_diagrams_avranks_percep_v_ens_gmean}.

These results can be seen as an indication that when scale matters to the ensemble, monolithic models may dictate the data scale-variability behavior of the ensembles built with them as their base model. It is, therefore, an important finding since, in order to choose the optimal scaling technique for an ensemble, tests can be performed with only one instance of the base model. This incurs significantly less computational cost than running the same tests with the entire ensemble.

Another valuable take from Figures \ref{fig:CD_diagrams_avranks_percep_v_ens_f1} and \ref{fig:CD_diagrams_avranks_percep_v_ens_gmean} is that, for all ensemble models, the performance with nonscaled data (NS) is statistically similar to at least one of the adjacent scaling techniques in the ranking, even for those ensembles that are more prone to variation due to scale, such as the Bagging model. Hence, choosing the wrong scaling technique can be as bad as not scaling the data at all.

\subsection{Scale-sensitivity vs. model performance}
\label{sec:mean_range_vs_avg_ranking}

If we look at both Figure \ref{fig:barplots_range} and Table \ref{tab:mean_performances} we can see that there are models that yield higher performances, such as KNORAE (close to 0.6 for F1 and 0.7 for G-Mean) and other models that are very sensitive to the choice of scaling technique, such as GP, that presents a mean range close to 0.26 for both metrics. However, it is hard to see from these resources how scale-sensitivity and model performance relate to each other.

\begin{figure*}[ht!]
    \centering
    \subfloat[\label{fig:mean_range_vs_avg_ranking_f1}]{
        \includegraphics[width=.68\textwidth]{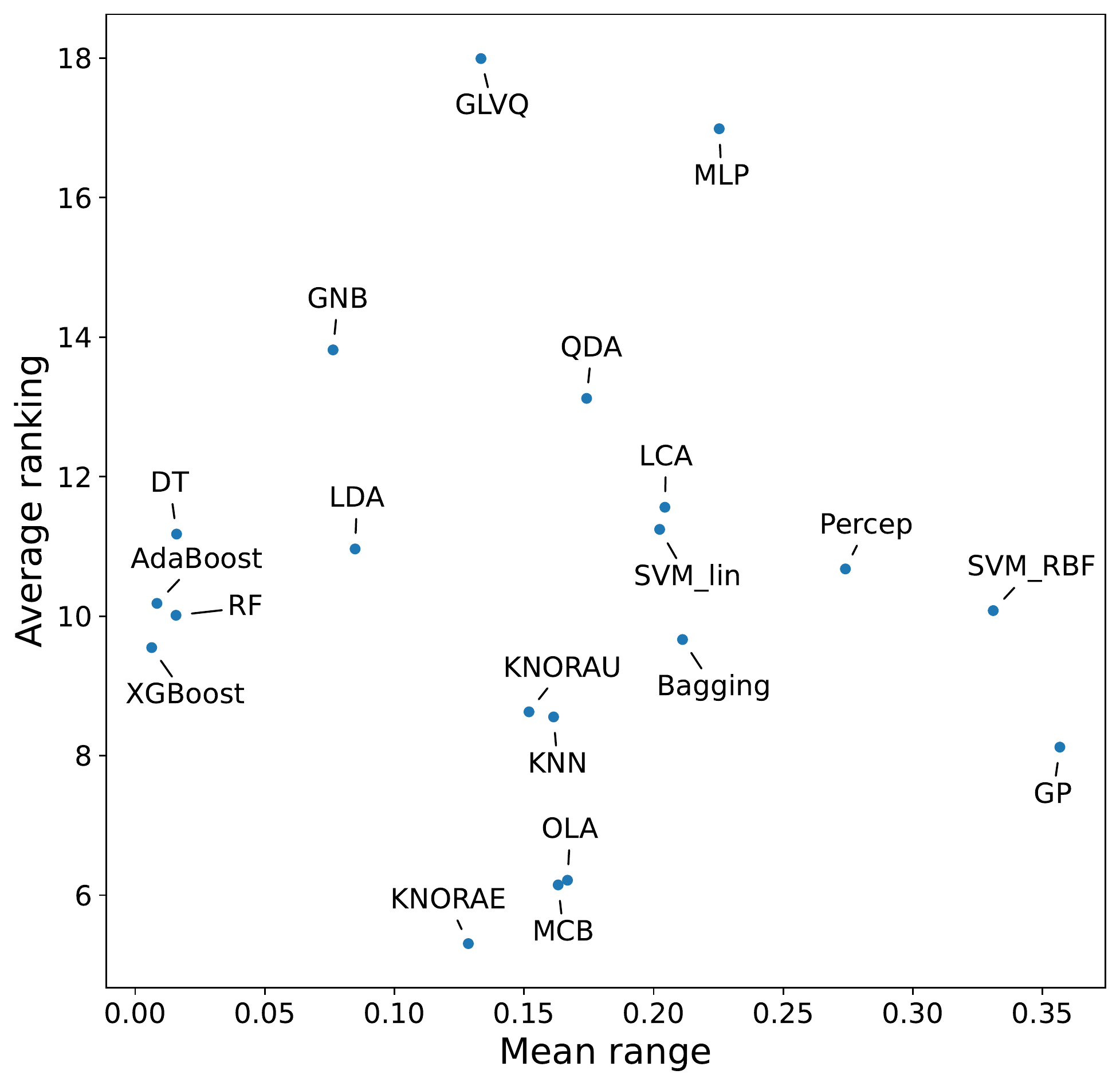}
    }
    
    \subfloat[\label{fig:mean_range_vs_avg_ranking_gmean}]{
        \includegraphics[width=.68\textwidth]{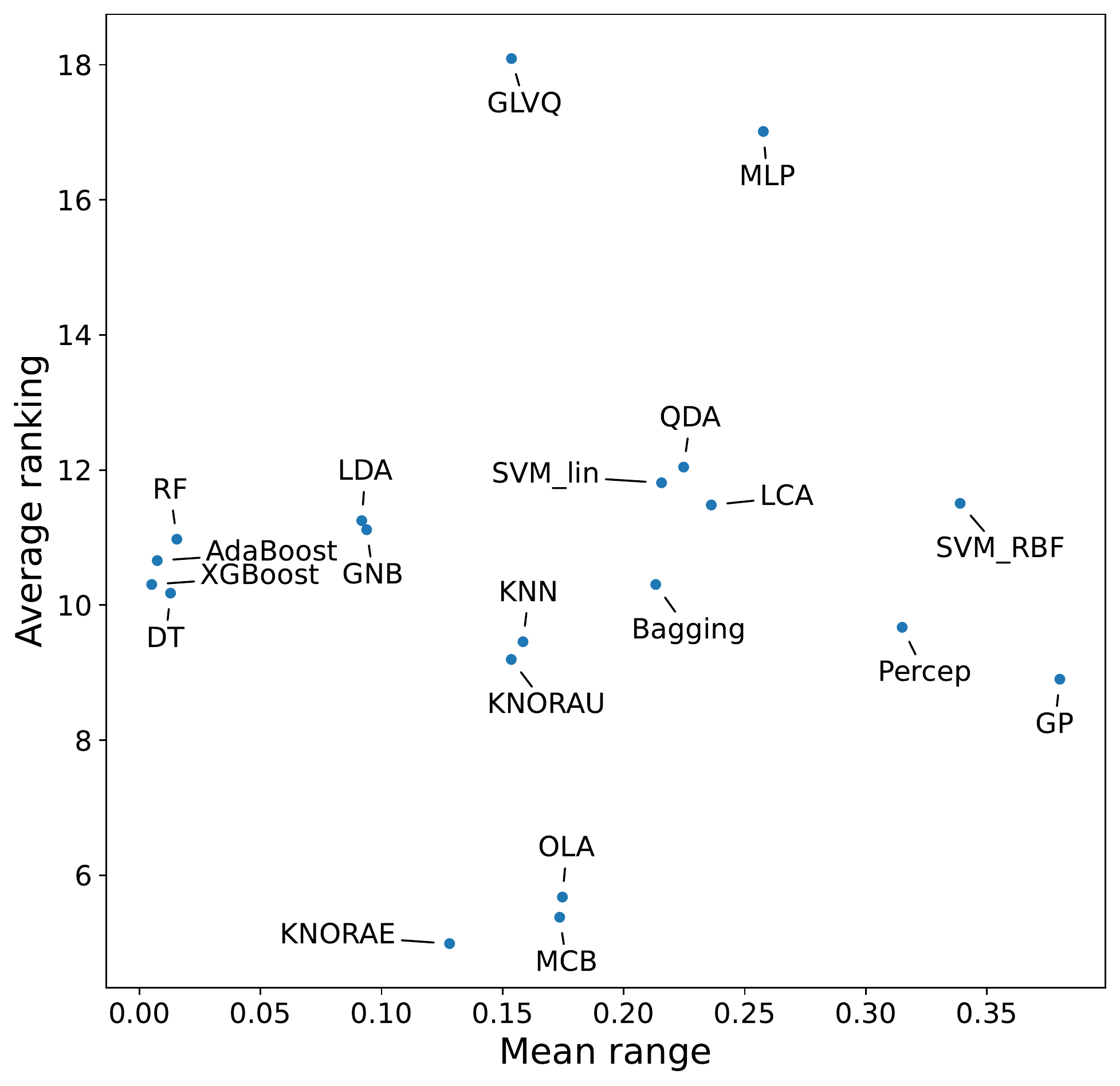}
    }
    \caption{Scatter plots relating model scale-sensitivity (mean range) with its performance (average ranking) considering metrics (a) F1 and (b) G-Mean.
    }
    \label{fig:mean_range_vs_avg_ranking}
\end{figure*}

In order to provide a better way to look at this relation, in Figure \ref{fig:mean_range_vs_avg_ranking} we present two scatter plots, one for each metric, where we compare the mean range against the average ranking of each model. For the mean range, we consider the ranges (differences between the performances of the best and the worst scaling techniques) obtained by a model over each of the 82 datasets and then calculate their mean. For the average ranking, we calculate the rank of a model compared to the others in each dataset and then we take the mean of all the 82 rankings of that model. Since each model is applied to six versions of the same dataset (one for each scaling technique), for the ranking, we take into account its best performance within the six versions. In these plots, the average ranking is better when it is lower. These scatter plots convey a way to analyze the models' scale-sensitivities and their performances, in terms of their average ranking over the 82 datasets.

This is an important analysis when one needs to select one model or a group of model for deployment in real-world applications with different requirements. For example, dataset scaling is often not an option for a real-time or online application in which model training is executed on the fly. Therefore, one would have to choose a model that performs well regardless of the data scale, such as DT or AdaBoost. In the other end of the spectrum, for a static and more conventional application, when training is done prior to deployment, it makes more sense to choose performance over scale stability, which would lead to the choice of a model such as KNORAE, MCB or OLA.

Another interesting observation from Figure \ref{fig:mean_range_vs_avg_ranking} is that the DT model and the DT based ensembles (AdaBoost, RF and XGBoost) appear in a cluster positioned in the middle left portion of the space. This means they are virtually insensitive to scale -- agreeing with previous discussions -- while presenting a reasonable performance. Additionally, the fact that they are together in a cluster means that there is little difference in choosing one over the others. This is somewhat surprising since we expected that these ensembles, even though static, would present a consistently better ranking throughout the datasets when compared to its base model.

On the other hand, the perceptron model (Percep) and the ensembles built with it as their base model (OLA, LCA, MCB, KNORAE and KNORAU) are more dispersely distributed in the space, with KNORAE being, at the same time, the best performant and the least sensitive to scale when compared to the other perceptron based ensembles. It also stands out that even though Percep and DT have very similar average ranking, their ensembles are positioned very differently in the space, which may be a consequence of their different nature, especially regarding their classifier selection method: static vs. dynamic. This could be further investigated by building dynamic ensembles of DTs as well as static ensembles of perceptrons. We expect that a DT-based KNORAE would be positioned closer to the origin in this graph.

\section{Related Works}
\label{sec:related}

This section chronologically presents a review of the few papers published so far, to the best of our knowledge, that present a comparative analysis of the effects of different scaling techniques on the classification performance of machine learning algorithms. In Table \ref{tab:related_works} we summarize the scaling techniques, the number of datasets, and the classification models that each of these papers have employed in their experiments. The last line presents the same information regarding the experiment conducted in this paper.

\input{tabs/tab_related_works}

We can see, from Table \ref{tab:related_works}, that the previous work lacked an analysis that is sufficiently diverse regarding all three perspectives: number of techniques, datasets and classification algorithms. 

Jain et al. \cite{jain2018} and Dzierżak et al. \cite{dzierzak2019}, for example, only considered two scaling techniques in their experiments, which were Min-max and Z-score normalization. In the case of Jain et al. \cite{jain2018}, authors aim their study on the dynamical selection of one of these two techniques based on data complexity measures. Although this paper reports the performances obtained by the techniques, a comparison is only minimally performed since its not their main focus. Dzierżak et al. \cite{dzierzak2019}, assessed the influence of these two scaling techniques on the performance of a model to detect osteoporosis and osteopenia from a single dataset comprising of 290 features representing computerized tomography (CT) images of sections of patients' spine. Their results indicated a superiority of the Z-score method in improving classification performance. However they did not employ a hypothesis test to evaluate the statistical significance of the improvement over the Min-max method.

While Raju et al. \cite{raju2020}, Singh et al. \cite{singh2020a} and Mishkov et al. \cite{Mishkov2022} all were able to compare significantly larger and more diverse sets of scaling techniques, they all failed to include a diverse set of classification algorithms. Additionally, when it comes to the generality of their findings, only in Singh et al. \cite{singh2020a} a relevant number of datasets was considered.

Raju et al. \cite{raju2020} applied seven distinct scaling techniques on a Kaggle diabetes dataset and then compared the performance of three classification models: KNN and two SVM variants. The authors compared the model's results with each scaling technique to the performance attained on original (nonscaled) data. These comparisons showed that scaling allowed a performance increase in every case, ranging from 5\% to 10\%. Endorsing our findings on the insensitivity of the KNN model to the choice of scaling technique, in their results KNN was the least sensitive when compared to the two SVM variants. In addition to compare the performance obtained with each scaling technique versus that of nonscaled data, it was also performed a comparison of the scaling techniques to one another. However, the paper lacked a more thorough analysis of the results along with hypothesis testing.

Singh et al. \cite{singh2020a} employed 14 scaling techniques in 21 datasets and then compared the performance of a k-Nearest Neighbors (KNN) classifier trained on the scaled and nonscaled datasets. The outcomes of the experiments when using the full feature set suggest that most of the methods help improve accuracy, nonscaled data does not always lead to the worst accuracy, which supports our conclusions in Section \ref{sec:answ_rq1_rq2}. Hypothesis tests confirmed that different techniques allow for significantly different results, but only two techniques (Pareto Scaling and Power Transformation) were significantly better than nonscaled data. However, the authors did not perform an analysis involving classification algorithms other than the KNN.

Finally, Mishkov et al. \cite{Mishkov2022} applied 16 scaling techniques to 4 different datasets in order to compare the classification accuracy of  a Multi-layer perceptron (MLP). Their results showed that the choice of scaling technique influences classification accuracy and suggested that this decision must be made on a case-by-case basis, i.e. depending on the dataset being analyzed. They also observed that nonscaled data do not always lead to the worst performance, agreeing with our findings in Section \ref{sec:answ_rq1_rq2}. This paper also lacked an analysis involving other classification algorithms.

To summarize, two of the five papers focus their analysis on just two scaling techniques \cite{jain2018, dzierzak2019}, while those that experiment with a more extensive number of techniques restrict their tests to just one  \cite{singh2020a, Mishkov2022}, or three classification algorithms \cite{raju2020}. Also, only in \cite{jain2018} and \cite{singh2020a}, authors provided results from a relevant number of datasets. Therefore, the literature lacks research works that consider various scaling techniques, a reasonable number of datasets, and classification algorithms simultaneously. Additionally, we notice that none of these works compared the effect of the scaling techniques in datasets affected by different imbalance ratios, neither measured their impact in ensemble models apart from the traditional Random Forest included in only two of the five papers.

It is also important to mention that, although in the two most recent papers \cite{singh2020a, Mishkov2022} authors employed a significantly larger set of scaling techniques than all the other works (including ours), some of those techniques are too similar or even equivalent to one another. As reported in \cite{singh2020a}, many of the techniques frequently return the same results for most datasets. Additionally, in \cite{Mishkov2022}, the set of techniques is increased largely to the decision of adding up to 7 variants of the same technique. In our paper, we aimed at selecting a diverse but concise set of techniques that we believe are representative enough of the scaling techniques available, as we detail in Section \ref{sec:scaling_tech}.

\section{Lessons learned}
\label{sec:lessons}

\begin{itemize}
    \item The choice of scaling technique matters for classification performance. Except for models that are, by nature, not sensitive to scale, the performance difference between the best and the worst scaling technique is relevant and statistically significant in most cases.
    \item Choosing the wrong scaling technique can be more detrimental to the classification performance than not scaling the data at all. For the SVM\_lin model, for example, the difference in F1 was higher than 0.5 for 32\% of the datasets, and higher than 0.8 for 7\% of them.
    \item The difference in performance considering the distinct scaling techniques can be observed across datasets presenting all levels of class imbalance ratio. However, it seems to be more salient for datasets where IR is higher.
    \item The Standard Scaler technique was the most successful on low IR datasets but the Quantile Transformer technique was better for all other IR strata. It indicates that IR may play an important role on the selection of the best scaling technique for a dataset.
    \item Models such as SVMs, GLVQ, Gaussian Process, Perceptron, MLP, Bagging, OLA, LCA and KNORA-U are very sensitive to the choice of scaling technique.
    \item Decision Trees (DT) and DT-based ensembles are virtually insensitive to the choice of scaling technique. On the other hand, ensembles built with Perceptrons present a wider range of sensitivity to the choice of scaling techniques and varying F1 and G-Mean performances.
    \item For ensembles that are more sensitive to the choice of scaling techniques, such as Bagging, LCA and KNORAU, the scaling techniques tend to rank similarly to their base models. This indicates that, when scale matters to the ensemble, base models may dictate the data scale-variability behavior of their ensembles. Therefore, testing which scaling technique is the best for an ensemble can be done by running only one instance of its base model, incurring in significantly lower computational cost.
    \item For real-time or online classification applications, in which model training is executed on the fly, with the data streaming into the learning process and therefore are hard to be effectively scaled, models such as DT or AdaBoost are preferable since they present the least sensitivity to scale while attaining reasonable performances.
    \item For static applications, when one can choose performance over scale stability, KNORAE, MCB or OLA are the most reasonable choices since they are the ones that present the best mean performances given that data is adequately scaled.
       
\end{itemize}

\section{Conclusion}
\label{sec:conclusion}

This study sought to understand whether the choice of the scaling technique significantly influences the performance of classification models and whether this influence changes according to the types of models used, including monolithic and ensemble models. We performed a broad experiment with five scaling techniques, 82 binary datasets, presenting a wide range of imbalance ratios, and 20 classification models, comprising 11 monolithic algorithms from five different subcategories and nine ensemble algorithms, including static, DCS and DES models.

Results indicated that the choice of scaling technique significantly affects classification performance and that choosing the wrong scaling technique can even be worse than not scaling the data at all. 
The analysis of scale sensitivity versus model performance  revealed the relationship between an ensemble and its base models when we look at their performance variations. It also showed that there is no ``one size fits all'' when it comes to model deployment, where application constraints, such as the feasibility of scaling and other requirements may determine the more appropriate model. This endorses the importance of such an analysis as a tool for model selection.

In future works, we intend to further investigate this relation between the performance variability due to the choice of scaling technique of an ensemble and its base model, e.g. better understand how an ensemble, when compared to its base model, is able to reduce the performance range considering different scaling techniques and how different configurations and types of ensembles present themselves in the graph in Figure \ref{fig:mean_range_vs_avg_ranking}. Another direction can be to understand which data characteristics are important to the selection of scaling techniques, which can also subsidize meta-learning research aiming at the dynamic selection of scaling techniques.

\bibliography{references}

\end{document}

%% file: tabs/tab_params.tex
\begin{table*}[ht]
\centering
\caption{Parameters used to build the classification models. Other unlisted parameters were left as default according to their respective libraries. DCS - Dynamic Classifier Selection, DES - Dynamic Ensemble Selection. The last five models inherited their pools from the Bagging static model listed above them.}
\label{tab:params}
\resizebox{0.95\textwidth}{!}{%
\begin{tabular}{clllll}
\hline
\textbf{Cat.}                                                                  & \textbf{Subcat.}                                                                  & \multicolumn{1}{c}{\textbf{Name}}                                                                           & \multicolumn{1}{c}{\textbf{Model}} & \multicolumn{1}{c}{\textbf{Parameters}}                                                                                                    & \multicolumn{1}{c}{\textbf{Library}}        \\ \hline
                                                                               &                                                                                   & Support Vector  Machine                                                                                     & SVM\_lin                           & kernel = 'linear'                                                                                                                          & sklearn v0.23.1                             \\
                                                                               &                                                                                   & \cellcolor[HTML]{EFEFEF}Support Vector  Machine                                                             & \cellcolor[HTML]{EFEFEF}SVM\_RBF   & \cellcolor[HTML]{EFEFEF}kernel = 'rbf'                                                                                                     & \cellcolor[HTML]{EFEFEF}sklearn v0.23.1     \\
                                                                               &                                                                                   & k-Nearest Neighbors                                                                                         & KNN                                & n\_neighbors = 5, n\_jobs=-1                                                                                                               & sklearn v0.23.1                             \\
                                                                               & \multirow{-4}{*}{\begin{tabular}[c]{@{}l@{}}Instance-\\ based\end{tabular}}       & \cellcolor[HTML]{EFEFEF}\begin{tabular}[c]{@{}l@{}}Generalized Learning \\ Vector Quantization\end{tabular} & \cellcolor[HTML]{EFEFEF}GLVQ       & \cellcolor[HTML]{EFEFEF}\begin{tabular}[c]{@{}l@{}}prototypes\_per\_class=1, max\_iter=2500, \\ gtol=1e-05, beta=2\end{tabular}            & \cellcolor[HTML]{EFEFEF}sklearn\_lvq v1.1.0 \\ \cline{2-6} 
                                                                               &                                                                                   & Gaussian Naive Bayes                                                                                        & GNB                                & --                                                                                                                                         & sklearn v0.23.1                             \\
                                                                               & \multirow{-2}{*}{\begin{tabular}[c]{@{}l@{}}Probabi-\\ listic\end{tabular}}       & \cellcolor[HTML]{EFEFEF}Gaussian Process                                                                    & \cellcolor[HTML]{EFEFEF}GP         & \cellcolor[HTML]{EFEFEF}kernel = 1.0 RBF(1.0),                                                                                             & \cellcolor[HTML]{EFEFEF}sklearn v0.23.1     \\ \cline{2-6} 
                                                                               &                                                                                   & \begin{tabular}[c]{@{}l@{}}Linear Discriminant \\ Analysis\end{tabular}                                     & LDA                                & solver='svd', tol=0.0001                                                                                                                   & sklearn v0.23.1                             \\
                                                                               & \multirow{-2}{*}{\begin{tabular}[c]{@{}l@{}}Discriminant\\ analysis\end{tabular}} & \cellcolor[HTML]{EFEFEF}\begin{tabular}[c]{@{}l@{}}Quadratic \\ Discriminant Analysis\end{tabular}         & \cellcolor[HTML]{EFEFEF}QDA        & \cellcolor[HTML]{EFEFEF}tol=0.0001                                                                                                         & \cellcolor[HTML]{EFEFEF}sklearn v0.23.1     \\ \cline{2-6} 
                                                                               & Rule-based                                                                        & Decision Tree                                                                                               & DT                                 & criterion='gini', splitter='best'                                                                                                          & sklearn v0.23.1                             \\ \cline{2-6} 
                                                                               &                                                                                   & \cellcolor[HTML]{EFEFEF}Perceptron                                                                          & \cellcolor[HTML]{EFEFEF}Percep     & \cellcolor[HTML]{EFEFEF}alpha=0.0001, max\_iter=1000, tol=0.001                                                                            & \cellcolor[HTML]{EFEFEF}sklearn v0.23.1     \\
\multirow{-11}{*}{\rotatebox[origin=c]{90}{Monolithic}} & \multirow{-2}{*}{\begin{tabular}[c]{@{}l@{}}Neural\\Networks\end{tabular}}         & Multi-layer Perceptron                                                                                      & MLP                                & \begin{tabular}[c]{@{}l@{}}activation='relu', solver='adam', alpha=1e-5, \\ hidden\_layer\_sizes=(5, 2)\end{tabular}                       & sklearn v0.23.1                             \\ \hline
                                                                               &                                                                                   & \cellcolor[HTML]{EFEFEF}\begin{tabular}[c]{@{}l@{}}eXtreme Gradient\\ Boosting\end{tabular}                 & \cellcolor[HTML]{EFEFEF}XGBoost    & \cellcolor[HTML]{EFEFEF}\begin{tabular}[c]{@{}l@{}}n\_estimators=100, importance\_type='gain',\\  tree\_method=auto\end{tabular}           & \cellcolor[HTML]{EFEFEF}xgboost 1.2.1       \\
                                                                               &                                                                                   & Random Forests                                                                                               & RF                                 & n\_estimators=100, criterion='gini'                                                                                                        & sklearn v0.23.1                             \\
                                                                               &                                                                                   & \cellcolor[HTML]{EFEFEF}AdaBoost                                                                            & \cellcolor[HTML]{EFEFEF}AdaBoost   & \cellcolor[HTML]{EFEFEF}\begin{tabular}[c]{@{}l@{}}n\_estimators=100, base\_estimator=\\ DecisionTreeClassifier(max\_depth=1)\end{tabular} & \cellcolor[HTML]{EFEFEF}sklearn v0.23.1     \\
                                                                               & \multirow{-4}{*}{Static}                                                          & Bagging                                                                                                     & Bagging                            & n\_estimators=100, base\_estimator=Perceptron                                                                                              & sklearn v0.23.1                             \\ \cline{2-6} 
                                                                               &                                                                                   & \cellcolor[HTML]{EFEFEF}\begin{tabular}[c]{@{}l@{}}Overall Local \\ Accuracy (OLA)\end{tabular}        & \cellcolor[HTML]{EFEFEF}OLA        & \cellcolor[HTML]{EFEFEF}pool\_classifiers={[}Bagging pool{]}                                                                               & \cellcolor[HTML]{EFEFEF}deslib v0.3.5       \\
                                                                               &                                                                                   & \begin{tabular}[c]{@{}l@{}}Local Class \\ Accuracy (LCA)\end{tabular}                                       & LCA                                & pool\_classifiers={[}Bagging pool{]}                                                                                                       & deslib v0.3.5                               \\
                                                                               & \multirow{-3}{*}{DCS}                                                             & \cellcolor[HTML]{EFEFEF}\begin{tabular}[c]{@{}l@{}}Multiple Classifier \\ Behaviour (MCB)\end{tabular}      & \cellcolor[HTML]{EFEFEF}MCB        & \cellcolor[HTML]{EFEFEF}pool\_classifiers={[}Bagging pool{]}                                                                               & \cellcolor[HTML]{EFEFEF}deslib v0.3.5       \\ \cline{2-6} 
                                                                               &                                                                                   & \begin{tabular}[c]{@{}l@{}}k-Nearest Oracles \\ Eliminate (KNORA-E)\end{tabular}                            & KNORAE                             & pool\_classifiers={[}Bagging pool{]},  k=7                                                                                                 & deslib v0.3.5                               \\
\multirow{-9}{*}{\rotatebox[origin=c]{90}{Ensemble}}    & \multirow{-2}{*}{DES}                                                             & \cellcolor[HTML]{EFEFEF}\begin{tabular}[c]{@{}l@{}}k-Nearest Oracles \\ Union (KNORA-U)\end{tabular}        & \cellcolor[HTML]{EFEFEF}KNORAU     & \cellcolor[HTML]{EFEFEF}pool\_classifiers={[}Bagging pool{]},  k=7                                                                         & \cellcolor[HTML]{EFEFEF}deslib v0.3.5       \\ \hline
\end{tabular}%
}
\vspace{-4mm} 
\end{table*}

%% file: tabs/tab_descDS2.tex
\begin{table*}[ht]
\centering
\caption{Datasets description. Note: Num. attrib. is the number of numerical attributes, while Categ. attrib. is the number of categorical attributes.}
\label{tab:descDS}
\resizebox{0.495\textwidth}{!}{%
\begin{tabular}{|llllll|}
\hline
\multicolumn{1}{c}{\textbf{\#}} & \multicolumn{1}{c}{\textbf{\begin{tabular}[c]{@{}c@{}}Dataset \\ name\end{tabular}}} & \multicolumn{1}{c}{\textbf{\begin{tabular}[c]{@{}c@{}}Num.\\ attrib.\end{tabular}}} & \multicolumn{1}{c}{\textbf{\begin{tabular}[c]{@{}c@{}}Categ.\\ attrib.\end{tabular}}} & \multicolumn{1}{c}{\textbf{\begin{tabular}[c]{@{}c@{}}Class\\ counts \end{tabular}}} & \multicolumn{1}{c}{\textbf{IR}} \\ \hline
\rowcolor[HTML]{EBEBEB} 
1                               & \cellcolor[HTML]{EBEBEB}glass1                                                                   & 9                                                                                          & 0                                                                                         & (138, 76)                                                                                   & 1.82                            \\
2                               & ecoli-0\_vs\_1                                                                                   & 7                                                                                          & 0                                                                                         & (143, 77)                                                                                   & 1.86                            \\
\rowcolor[HTML]{EBEBEB} 
3                               & wisconsin                                                                                        & 9                                                                                          & 0                                                                                         & (444, 239)                                                                                  & 1.86                            \\
4                               & pima                                                                                             & 8                                                                                          & 0                                                                                         & (268, 500)                                                                                  & 1.87                            \\
\rowcolor[HTML]{EBEBEB} 
5                               & iris0                                                                                            & 4                                                                                          & 0                                                                                         & (50, 100)                                                                                   & 2.00                            \\
6                               & glass0                                                                                           & 9                                                                                          & 0                                                                                         & (70, 144)                                                                                   & 2.06                            \\
\rowcolor[HTML]{EBEBEB} 
7                               & yeast1                                                                                           & 8                                                                                          & 0                                                                                         & (1055, 429)                                                                                 & 2.46                            \\
8                               & haberman                                                                                         & 3                                                                                          & 0                                                                                         & (225, 81)                                                                                   & 2.78                            \\
\rowcolor[HTML]{EBEBEB} 
9                               & vehicle2                                                                                         & 18                                                                                         & 0                                                                                         & (628, 218)                                                                                  & 2.88                            \\
10                              & vehicle1                                                                                         & 18                                                                                         & 0                                                                                         & (629, 217)                                                                                  & 2.90                            \\
\rowcolor[HTML]{EBEBEB} 
11                              & vehicle3                                                                                         & 18                                                                                         & 0                                                                                         & (634, 212)                                                                                  & 2.99                            \\
12                              & glass-0-1-2-3\_vs\_4-5-6                                                                         & 9                                                                                          & 0                                                                                         & (163, 51)                                                                                   & 3.20                            \\
\rowcolor[HTML]{EBEBEB} 
13                              & vehicle0                                                                                         & 18                                                                                         & 0                                                                                         & (199, 647)                                                                                  & 3.25                            \\
14                              & ecoli1                                                                                           & 7                                                                                          & 0                                                                                         & (259, 77)                                                                                   & 3.36                            \\
\rowcolor[HTML]{EBEBEB} 
15                              & new-thyroid1                                                                                     & 5                                                                                          & 0                                                                                         & (180, 35)                                                                                   & 5.14                            \\
16                              & ecoli2                                                                                           & 7                                                                                          & 0                                                                                         & (284, 52)                                                                                   & 5.46                            \\
\rowcolor[HTML]{EBEBEB} 
17                              & segment0                                                                                         & 19                                                                                         & 0                                                                                         & (1979, 329)                                                                                 & 6.02                            \\
18                              & glass6                                                                                           & 9                                                                                          & 0                                                                                         & (185, 29)                                                                                   & 6.38                            \\
\rowcolor[HTML]{EBEBEB} 
19                              & yeast3                                                                                           & 8                                                                                          & 0                                                                                         & (1321, 163)                                                                                 & 8.10                            \\
20                              & ecoli3                                                                                           & 7                                                                                          & 0                                                                                         & (301, 35)                                                                                   & 8.60                            \\
\rowcolor[HTML]{EBEBEB} 
21                              & page-blocks0                                                                                     & 10                                                                                         & 0                                                                                         & (4913, 559)                                                                                 & 8.79                            \\
22                              & ecoli-0-3-4\_vs\_5                                                                               & 7                                                                                          & 0                                                                                         & (180, 20)                                                                                   & 9.00                            \\
\rowcolor[HTML]{EBEBEB} 
23                              & yeast-2\_vs\_4                                                                                   & 8                                                                                          & 0                                                                                         & (463, 51)                                                                                   & 9.08                            \\
24                              & ecoli-0-6-7\_vs\_3-5                                                                             & 7                                                                                          & 0                                                                                         & (200, 22)                                                                                   & 9.09                            \\
\rowcolor[HTML]{EBEBEB} 
25                              & ecoli-0-2-3-4\_vs\_5                                                                             & 7                                                                                          & 0                                                                                         & (182, 20)                                                                                   & 9.10                            \\
26                              & glass-0-1-5\_vs\_2                                                                               & 9                                                                                          & 0                                                                                         & (155, 17)                                                                                   & 9.12                            \\
\rowcolor[HTML]{EBEBEB} 
27                              & yeast-0-3-5-9\_vs\_7-8                                                                           & 8                                                                                          & 0                                                                                         & (456, 50)                                                                                   & 9.12                            \\
28                              & yeast-0-2-5-6\_vs\_3-7-8-9                                                                       & 8                                                                                          & 0                                                                                         & (905, 99)                                                                                   & 9.14                            \\
\rowcolor[HTML]{EBEBEB} 
29                              & ecoli-0-4-6\_vs\_5                                                                               & 6                                                                                          & 0                                                                                         & (183, 20)                                                                                   & 9.15                            \\
30                              & ecoli-0-1\_vs\_2-3-5                                                                             & 7                                                                                          & 0                                                                                         & (220, 24)                                                                                   & 9.17                            \\
\rowcolor[HTML]{EBEBEB} 
31                              & ecoli-0-2-6-7\_vs\_3-5                                                                           & 7                                                                                          & 0                                                                                         & (202, 22)                                                                                   & 9.18                            \\
32                              & glass-0-4\_vs\_5                                                                                 & 9                                                                                          & 0                                                                                         & (83, 9)                                                                                     & 9.22                            \\
\rowcolor[HTML]{EBEBEB} 
33                              & ecoli-0-3-4-6\_vs\_5                                                                             & 7                                                                                          & 0                                                                                         & (185, 20)                                                                                   & 9.25                            \\
34                              & ecoli-0-3-4-7\_vs\_5-6                                                                           & 7                                                                                          & 0                                                                                         & (232, 25)                                                                                   & 9.28                            \\
\rowcolor[HTML]{EBEBEB} 
35                              & yeast-0-5-6-7-9\_vs\_4                                                                           & 8                                                                                          & 0                                                                                         & (477, 51)                                                                                   & 9.35                            \\
36                              & vowel0                                                                                           & 13                                                                                         & 0                                                                                         & (90, 898)                                                                                   & 9.98                            \\
\rowcolor[HTML]{EBEBEB} 
37                              & ecoli-0-6-7\_vs\_5                                                                               & 6                                                                                          & 0                                                                                         & (200, 20)                                                                                   & 10.00                           \\
38                              & glass-0-1-6\_vs\_2                                                                               & 9                                                                                          & 0                                                                                         & (175, 17)                                                                                   & 10.29                           \\
\rowcolor[HTML]{EBEBEB} 
39                              & ecoli-0-1-4-7\_vs\_2-3-5-6                                                                       & 7                                                                                          & 0                                                                                         & (307, 29)                                                                                   & 10.59                           \\
40                              & led7digit-0-2-4-5-6-7-8-9\_vs\_1                                                                 & 7                                                                                          & 0                                                                                         & (406, 37)                                                                                   & 10.97                           \\
\rowcolor[HTML]{EBEBEB} 
41                              & ecoli-0-1\_vs\_5                                                                                 & 6                                                                                          & 0                                                                                         & (220, 20)                                                                                   & 11.00                           \\ \hline
\end{tabular}%
}
\resizebox{0.492\textwidth}{!}{%
\begin{tabular}{|llllll|}
\hline
\multicolumn{1}{c}{\textbf{\#}} & \multicolumn{1}{c}{\textbf{\begin{tabular}[c]{@{}c@{}}Dataset \\ name\end{tabular}}} & \multicolumn{1}{c}{\textbf{\begin{tabular}[c]{@{}c@{}}Num.\\ attrib.\end{tabular}}} & \multicolumn{1}{c}{\textbf{\begin{tabular}[c]{@{}c@{}}Categ.\\ attrib.\end{tabular}}} & \multicolumn{1}{c}{\textbf{\begin{tabular}[c]{@{}c@{}}Class\\ counts \end{tabular}}} & \multicolumn{1}{c}{\textbf{IR}} \\ \hline
42                              & glass-0-1-4-6\_vs\_2                                                                             & 9                                                                                          & 0                                                                                         & (188, 17)                                                                                   & 11.06                           \\
\rowcolor[HTML]{EBEBEB} 
43                              & glass2                                                                                           & 9                                                                                          & 0                                                                                         & (197, 17)                                                                                   & 11.59                           \\
44                              & ecoli-0-1-4-7\_vs\_5-6                                                                           & 6                                                                                          & 0                                                                                         & (307, 25)                                                                                   & 12.28                           \\
\rowcolor[HTML]{EBEBEB} 
45                              & cleveland-0\_vs\_4                                                                               & 13                                                                                         & 0                                                                                         & (160, 13)                                                                                   & 12.31                           \\
46                              & ecoli-0-1-4-6\_vs\_5                                                                             & 6                                                                                          & 0                                                                                         & (260, 20)                                                                                   & 13.00                           \\
\rowcolor[HTML]{EBEBEB} 
47                              & shuttle-c0-vs-c4                                                                                 & 9                                                                                          & 0                                                                                         & (1706, 123)                                                                                 & 13.87                           \\
48                              & yeast-1\_vs\_7                                                                                   & 7                                                                                          & 0                                                                                         & (429, 30)                                                                                   & 14.30                           \\
\rowcolor[HTML]{EBEBEB} 
49                              & glass4                                                                                           & 9                                                                                          & 0                                                                                         & (201, 13)                                                                                   & 15.46                           \\
50                              & ecoli4                                                                                           & 7                                                                                          & 0                                                                                         & (316, 20)                                                                                   & 15.80                           \\
\rowcolor[HTML]{EBEBEB} 
51                              & page-blocks-1-3\_vs\_4                                                                           & 10                                                                                         & 0                                                                                         & (444, 28)                                                                                   & 15.86                           \\
52                              & abalone9-18                                                                                      & 7                                                                                          & 1                                                                                         & (689, 42)                                                                                   & 16.40                           \\
\rowcolor[HTML]{EBEBEB} 
53                              & dermatology-6                                                                                    & 34                                                                                         & 0                                                                                         & (338, 20)                                                                                   & 16.90                           \\
54                              & glass-0-1-6\_vs\_5                                                                               & 9                                                                                          & 0                                                                                         & (175, 9)                                                                                    & 19.44                           \\
\rowcolor[HTML]{EBEBEB} 
55                              & shuttle-c2-vs-c4                                                                                 & 9                                                                                          & 0                                                                                         & (6, 123)                                                                                    & 20.50                           \\
56                              & shuttle-6\_vs\_2-3                                                                               & 9                                                                                          & 0                                                                                         & (220, 10)                                                                                   & 22.00                           \\
\rowcolor[HTML]{EBEBEB} 
57                              & yeast-1-4-5-8\_vs\_7                                                                             & 8                                                                                          & 0                                                                                         & (663, 30)                                                                                   & 22.10                           \\
58                              & glass5                                                                                           & 9                                                                                          & 0                                                                                         & (205, 9)                                                                                    & 22.78                           \\
\rowcolor[HTML]{EBEBEB} 
59                              & yeast-2\_vs\_8                                                                                   & 8                                                                                          & 0                                                                                         & (462, 20)                                                                                   & 23.10                           \\
60                              & yeast4                                                                                           & 8                                                                                          & 0                                                                                         & (1433, 51)                                                                                  & 28.10                           \\
\rowcolor[HTML]{EBEBEB} 
61                              & winequality-red-4                                                                                & 11                                                                                         & 0                                                                                         & (1546, 53)                                                                                  & 29.17                           \\
62                              & poker-9\_vs\_7                                                                                   & 10                                                                                         & 0                                                                                         & (236, 8)                                                                                    & 29.50                           \\
\rowcolor[HTML]{EBEBEB} 
63                              & yeast-1-2-8-9\_vs\_7                                                                             & 8                                                                                          & 0                                                                                         & (917, 30)                                                                                   & 30.57                           \\
64                              & abalone-3\_vs\_11                                                                                & 7                                                                                          & 1                                                                                         & (15, 487)                                                                                   & 32.47                           \\
\rowcolor[HTML]{EBEBEB} 
65                              & winequality-white-9\_vs\_4                                                                       & 11                                                                                         & 0                                                                                         & (163, 5)                                                                                    & 32.60                           \\
66                              & yeast5                                                                                           & 8                                                                                          & 0                                                                                         & (1440, 44)                                                                                  & 32.73                           \\
\rowcolor[HTML]{EBEBEB} 
67                              & winequality-red-8\_vs\_6                                                                         & 11                                                                                         & 0                                                                                         & (638, 18)                                                                                   & 35.44                           \\
68                              & ecoli-0-1-3-7\_vs\_2-6                                                                           & 7                                                                                          & 0                                                                                         & (274, 7)                                                                                    & 39.14                           \\
\rowcolor[HTML]{EBEBEB} 
69                              & abalone-17\_vs\_7-8-9-10                                                                         & 7                                                                                          & 1                                                                                         & (2280, 58)                                                                                  & 39.31                           \\
70                              & abalone-21\_vs\_8                                                                                & 7                                                                                          & 1                                                                                         & (14, 567)                                                                                   & 40.50                           \\
\rowcolor[HTML]{EBEBEB} 
71                              & yeast6                                                                                           & 8                                                                                          & 0                                                                                         & (1449, 35)                                                                                  & 41.40                           \\
72                              & winequality-white-3\_vs\_7                                                                       & 11                                                                                         & 0                                                                                         & (880, 20)                                                                                   & 44.00                           \\
\rowcolor[HTML]{EBEBEB} 
73                              & winequality-red-8\_vs\_6-7                                                                       & 11                                                                                         & 0                                                                                         & (837, 18)                                                                                   & 46.50                           \\
74                              & abalone-19\_vs\_10-11-12-13                                                                      & 7                                                                                          & 1                                                                                         & (32, 1590)                                                                                  & 49.69                           \\
\rowcolor[HTML]{EBEBEB} 
75                              & winequality-white-3-9\_vs\_5                                                                     & 11                                                                                         & 0                                                                                         & (1457, 25)                                                                                  & 58.28                           \\
76                              & poker-8-9\_vs\_6                                                                                 & 10                                                                                         & 0                                                                                         & (1460, 25)                                                                                  & 58.40                           \\
\rowcolor[HTML]{EBEBEB} 
77                              & shuttle-2\_vs\_5                                                                                 & 9                                                                                          & 0                                                                                         & (3267, 49)                                                                                  & 66.67                           \\
78                              & winequality-red-3\_vs\_5                                                                         & 11                                                                                         & 0                                                                                         & (681, 10)                                                                                   & 68.10                           \\
\rowcolor[HTML]{EBEBEB} 
79                              & abalone-20\_vs\_8-9-10                                                                           & 7                                                                                          & 1                                                                                         & (1890, 26)                                                                                  & 72.69                           \\
80                              & poker-8-9\_vs\_5                                                                                 & 10                                                                                         & 0                                                                                         & (2050, 25)                                                                                  & 82.00                           \\
\rowcolor[HTML]{EBEBEB} 
81                              & poker-8\_vs\_6                                                                                   & 10                                                                                         & 0                                                                                         & (1460, 17)                                                                                  & 85.88                           \\
82                              & abalone19                                                                                        & 7                                                                                          & 1                                                                                         & (4142, 32)                                                                                  & 129.44                          \\ \hline
\end{tabular}

}
\end{table*}

%% file: tabs/tab_mean_performances.tex
\begin{table*}[!ht]
\centering
\caption{Mean performances of the classification models. The table shows, for each model, the mean obtained with the pair of metric and scaling technique. Each value is a mean calculated over the 82 datasets.}
\label{tab:mean_performances}
\resizebox{0.9\textwidth}{!}{%
\begin{tabular}{l|llllll|llllll}
\hline
                                 & \multicolumn{6}{c|}{\textbf{F1}}                                                                                                                                                        & \multicolumn{6}{c}{\textbf{G-Mean}}                                                                                                                                                     \\ \cline{2-13} 
\multirow{-2}{*}{\textbf{Model}} & \multicolumn{1}{c}{\textbf{NS}} & \multicolumn{1}{c}{\textbf{SS}} & \multicolumn{1}{c}{\textbf{MM}} & \multicolumn{1}{c}{\textbf{MA}} & \multicolumn{1}{c}{\textbf{RS}} & \textbf{QT}   & \multicolumn{1}{c}{\textbf{NS}} & \multicolumn{1}{c}{\textbf{SS}} & \multicolumn{1}{c}{\textbf{MM}} & \multicolumn{1}{c}{\textbf{MA}} & \multicolumn{1}{c}{\textbf{RS}} & \textbf{QT}   \\ \hline
\rowcolor[HTML]{EFEFEF} 
SVM\_lin                         & 0.44                            & 0.51                            & 0.38                            & 0.36                            & 0.49                            & \textbf{0.52} & 0.48                            & 0.56                            & 0.41                            & 0.38                            & 0.54                            & \textbf{0.57} \\
SVM\_RBF                         & 0.35                            & 0.52                            & \textbf{0.48}                   & 0.45                            & 0.40                            & 0.45          & 0.38                            & \textbf{0.56}                   & 0.51                            & 0.48                            & 0.44                            & 0.49          \\
\rowcolor[HTML]{EFEFEF} 
KNN                              & 0.51                            & 0.55                            & 0.54                            & 0.53                            & 0.54                            & \textbf{0.56} & 0.56                            & 0.60                            & 0.58                            & 0.57                            & 0.59                            & \textbf{0.60} \\
GNB                              & 0.42                            & 0.40                            & 0.41                            & 0.41                            & 0.42                            & \textbf{0.42} & 0.61                            & 0.59                            & 0.60                            & 0.60                            & \textbf{0.62}                   & 0.57          \\
\rowcolor[HTML]{EFEFEF} 
GLVQ                             & 0.10                            & \textbf{0.11}                   & 0.08                            & 0.09                            & 0.17                            & 0.09          & 0.11                            & \textbf{0.11}                   & 0.08                            & 0.09                            & 0.19                            & 0.10          \\
LDA                              & \textbf{0.54}                   & \textbf{0.54}                   & \textbf{0.54}                   & \textbf{0.54}                   & \textbf{0.54}                   & 0.52          & \textbf{0.61}                   & \textbf{0.61}                   & \textbf{0.61}                   & \textbf{0.61}                   & \textbf{0.61}                   & 0.59          \\
\rowcolor[HTML]{EFEFEF} 
QDA                              & 0.36                            & 0.37                            & 0.33                            & 0.37                            & \textbf{0.38}                   & 0.36          & 0.47                            & 0.48                            & 0.45                            & 0.48                            & \textbf{0.49}                   & 0.43          \\
GP                               & 0.52                            & 0.36                            & \textbf{0.57}                   & 0.56                            & 0.44                            & 0.32          & 0.56                            & 0.38                            & \textbf{0.62}                   & 0.61                            & 0.48                            & 0.35          \\
\rowcolor[HTML]{EFEFEF} 
DT                               & 0.56                            & 0.56                            & 0.56                            & 0.56                            & \textbf{0.57}                   & 0.57          & 0.68                            & 0.68                            & 0.68                            & 0.68                            & \textbf{0.69}                   & 0.68          \\
Percep                           & 0.37                            & \textbf{0.52}                   & 0.47                            & 0.44                            & 0.48                            & 0.47          & 0.44                            & \textbf{0.63}                   & 0.55                            & 0.52                            & 0.60                            & 0.58          \\
\rowcolor[HTML]{EFEFEF} 
MLP                              & 0.08                            & 0.19                            & 0.05                            & 0.05                            & 0.18                            & \textbf{0.21} & 0.09                            & 0.21                            & 0.05                            & 0.05                            & 0.20                            & \textbf{0.23} \\
RF                               & 0.56                            & \textbf{0.56}                   & 0.55                            & 0.55                            & 0.56                            & 0.55          & 0.60                            & 0.60                            & 0.60                            & 0.60                            & \textbf{0.60}                   & 0.60          \\
\rowcolor[HTML]{EFEFEF} 
XGBoost                          & 0.59                            & 0.58                            & 0.59                            & \textbf{0.59}                   & 0.59                            & 0.58          & \textbf{0.65}                   & 0.65                            & 0.65                            & 0.65                            & 0.65                            & 0.64          \\
AdaBoost                         & 0.57                            & 0.57                            & 0.57                            & 0.57                            & 0.57                            & \textbf{0.57} & 0.65                            & 0.65                            & 0.64                            & 0.64                            & 0.64                            & \textbf{0.65} \\
\rowcolor[HTML]{EFEFEF} 
Bagging                          & 0.39                            & \textbf{0.56}                   & 0.50                            & 0.48                            & 0.52                            & 0.53          & 0.44                            & \textbf{0.62}                   & 0.55                            & 0.52                            & 0.59                            & 0.59          \\
OLA                              & 0.53                            & \textbf{0.60}                   & 0.60                            & 0.58                            & 0.58                            & 0.60          & 0.60                            & 0.70                            & 0.67                            & 0.65                            & 0.68                            & \textbf{0.70} \\
\rowcolor[HTML]{EFEFEF} 
LCA                              & 0.40                            & \textbf{0.49}                   & 0.46                            & 0.42                            & 0.47                            & 0.48          & 0.45                            & \textbf{0.56}                   & 0.53                            & 0.48                            & 0.54                            & 0.55          \\
MCB                              & 0.53                            & 0.59                            & \textbf{0.59}                   & 0.58                            & 0.59                            & 0.59          & 0.60                            & \textbf{0.69}                   & 0.67                            & 0.65                            & 0.68                            & 0.69          \\
\rowcolor[HTML]{EFEFEF} 
KNORAE                           & 0.58                            & 0.61                            & 0.61                            & 0.60                            & 0.61                            & \textbf{0.62} & 0.67                            & 0.70                            & 0.70                            & 0.69                            & 0.70                            & \textbf{0.71} \\
KNORAU                           & 0.48                            & \textbf{0.57}                   & 0.53                            & 0.50                            & 0.55                            & 0.55          & 0.53                            & \textbf{0.63}                   & 0.58                            & 0.55                            & 0.61                            & 0.61          \\ \hline
\end{tabular}%
}
\end{table*}

%% file: tabs/tab_friedman_diff_ST_low_med_high_all.tex
\begin{table*}[!ht]
\centering
\caption{Results for the Friedman hypothesis tests, each set of tests consider a stratum of datasets with different IR levels: low, medium, high and then, all the datasets. Check marks indicate p-values that reject the null hypothesis.}
\label{tab:friedman_diff_ST_low_med_high_all}
\resizebox{0.85\textwidth}{!}{%
\begin{tabular}{ll|cc|cc|cc|cc}
                                                                                      &                                        & \multicolumn{2}{c|}{\textbf{Low IR}}                       & \multicolumn{2}{c|}{\textbf{Medium IR}}                    & \multicolumn{2}{c|}{\textbf{High IR}}                      & \multicolumn{2}{c}{\textbf{All datasets}}                  \\ \cline{3-10} 
\multirow{-2}{*}{\textbf{Model}}                                                      & \multirow{-2}{*}{\textbf{Metric}}      & \multicolumn{2}{c|}{\textbf{p-value}}                      & \multicolumn{2}{c|}{\textbf{p-value}}                      & \multicolumn{2}{c|}{\textbf{p-value}}                      & \multicolumn{2}{c}{\textbf{p-value}}                       \\ \hline
                                                                                      & \cellcolor[HTML]{EFEFEF}F1             & \cellcolor[HTML]{EFEFEF}0.0040 & \cellcolor[HTML]{EFEFEF}\checkmark & \cellcolor[HTML]{EFEFEF}0.0121 & \cellcolor[HTML]{EFEFEF}\checkmark & \cellcolor[HTML]{EFEFEF}0.0000 & \cellcolor[HTML]{EFEFEF}\checkmark & \cellcolor[HTML]{EFEFEF}0.0000 & \cellcolor[HTML]{EFEFEF}\checkmark \\
\multirow{-2}{*}{\textbf{SVM\_lin}}                                                   & G-Mean                                 & 0.0017                         & \checkmark                         & 0.0004                         & \checkmark                         & 0.0000                         & \checkmark                         & 0.0000                         & \checkmark                         \\
                                                                                      & \cellcolor[HTML]{EFEFEF}F1             & \cellcolor[HTML]{EFEFEF}0.0082 & \cellcolor[HTML]{EFEFEF}\checkmark & \cellcolor[HTML]{EFEFEF}0.7154 & \cellcolor[HTML]{EFEFEF}  & \cellcolor[HTML]{EFEFEF}0.0001 & \cellcolor[HTML]{EFEFEF}\checkmark & \cellcolor[HTML]{EFEFEF}0.0000 & \cellcolor[HTML]{EFEFEF}\checkmark \\
\multirow{-2}{*}{\textbf{SVM\_RBF}}                                                   & G-Mean                                 & 0.0108                         & \checkmark                         & 0.5642                         &                           & 0.0001                         & \checkmark                         & 0.0000                         & \checkmark                         \\
                                                                                      & \cellcolor[HTML]{EFEFEF}F1             & \cellcolor[HTML]{EFEFEF}0.2732 & \cellcolor[HTML]{EFEFEF}  & \cellcolor[HTML]{EFEFEF}0.8664 & \cellcolor[HTML]{EFEFEF}  & \cellcolor[HTML]{EFEFEF}0.0722 & \cellcolor[HTML]{EFEFEF}  & \cellcolor[HTML]{EFEFEF}0.0182 & \cellcolor[HTML]{EFEFEF}\checkmark \\
\multirow{-2}{*}{\textbf{KNN}}                                                        & G-Mean                                 & 0.2500                         &                           & 0.9975                         &                           & 0.2172                         &                           & 0.0732                         &                           \\
                                                                                      & \cellcolor[HTML]{EFEFEF}F1             & \cellcolor[HTML]{EFEFEF}0.8886 & \cellcolor[HTML]{EFEFEF}  & \cellcolor[HTML]{EFEFEF}0.1820 & \cellcolor[HTML]{EFEFEF}  & \cellcolor[HTML]{EFEFEF}0.0001 & \cellcolor[HTML]{EFEFEF}\checkmark & \cellcolor[HTML]{EFEFEF}0.0000 & \cellcolor[HTML]{EFEFEF}\checkmark \\
\multirow{-2}{*}{\textbf{GNB}}                                                        & G-Mean                                 & 0.8886                         &                           & 0.0823                         &                           & 0.0000                         & \checkmark                         & 0.0000                         & \checkmark                         \\
                                                                                      & \cellcolor[HTML]{EFEFEF}F1             & \cellcolor[HTML]{EFEFEF}0.3480 & \cellcolor[HTML]{EFEFEF}  & \cellcolor[HTML]{EFEFEF}0.7904 & \cellcolor[HTML]{EFEFEF}  & \cellcolor[HTML]{EFEFEF}0.0000 & \cellcolor[HTML]{EFEFEF}\checkmark & \cellcolor[HTML]{EFEFEF}0.0027 & \cellcolor[HTML]{EFEFEF}\checkmark \\
\multirow{-2}{*}{\textbf{GLVQ}}                                                       & G-Mean                                 & 0.7364                         &                           & 0.7423                         &                           & 0.0000                         & \checkmark                         & 0.0039                         & \checkmark                         \\
                                                                                      & \cellcolor[HTML]{EFEFEF}F1             & \cellcolor[HTML]{EFEFEF}0.5364 & \cellcolor[HTML]{EFEFEF}  & \cellcolor[HTML]{EFEFEF}0.5364 & \cellcolor[HTML]{EFEFEF}  & \cellcolor[HTML]{EFEFEF}0.2239 & \cellcolor[HTML]{EFEFEF}  & \cellcolor[HTML]{EFEFEF}0.0132 & \cellcolor[HTML]{EFEFEF}\checkmark \\
\multirow{-2}{*}{\textbf{LDA}}                                                        & G-Mean                                 & 0.5364                         &                           & 0.0005                         & \checkmark                         & 0.2239                         &                           & 0.0002                         & \checkmark                         \\
                                                                                      & \cellcolor[HTML]{EFEFEF}F1             & \cellcolor[HTML]{EFEFEF}0.0016 & \cellcolor[HTML]{EFEFEF}\checkmark & \cellcolor[HTML]{EFEFEF}0.3045 & \cellcolor[HTML]{EFEFEF}  & \cellcolor[HTML]{EFEFEF}0.0512 & \cellcolor[HTML]{EFEFEF}  & \cellcolor[HTML]{EFEFEF}0.0052 & \cellcolor[HTML]{EFEFEF}\checkmark \\
\multirow{-2}{*}{\textbf{QDA}}                                                        & G-Mean                                 & 0.0078                         & \checkmark                         & 0.7324                         &                           & 0.0133                         & \checkmark                         & 0.0006                         & \checkmark                         \\
                                                                                      & \cellcolor[HTML]{EFEFEF}F1             & \cellcolor[HTML]{EFEFEF}0.0263 & \cellcolor[HTML]{EFEFEF}\checkmark & \cellcolor[HTML]{EFEFEF}0.2627 & \cellcolor[HTML]{EFEFEF}  & \cellcolor[HTML]{EFEFEF}0.0000 & \cellcolor[HTML]{EFEFEF}\checkmark & \cellcolor[HTML]{EFEFEF}0.0000 & \cellcolor[HTML]{EFEFEF}\checkmark \\
\multirow{-2}{*}{\textbf{GP}}                                                         & G-Mean                                 & 0.0405                         & \checkmark                         & 0.1699                         &                           & 0.0000                         & \checkmark                         & 0.0000                         & \checkmark                         \\
                                                                                      & \cellcolor[HTML]{EFEFEF}F1             & \cellcolor[HTML]{EFEFEF}0.5288 & \cellcolor[HTML]{EFEFEF}  & \cellcolor[HTML]{EFEFEF}0.0869 & \cellcolor[HTML]{EFEFEF}  & \cellcolor[HTML]{EFEFEF}0.0164 & \cellcolor[HTML]{EFEFEF}\checkmark & \cellcolor[HTML]{EFEFEF}0.1037 & \cellcolor[HTML]{EFEFEF}  \\
\multirow{-2}{*}{\textbf{DT}}                                                         & G-Mean                                 & 0.4136                         &                           & 0.1139                         &                           & 0.0361                         & \checkmark                         & 0.1198                         &                           \\
                                                                                      & \cellcolor[HTML]{EFEFEF}F1             & \cellcolor[HTML]{EFEFEF}0.0024 & \cellcolor[HTML]{EFEFEF}\checkmark & \cellcolor[HTML]{EFEFEF}0.0038 & \cellcolor[HTML]{EFEFEF}\checkmark & \cellcolor[HTML]{EFEFEF}0.0000 & \cellcolor[HTML]{EFEFEF}\checkmark & \cellcolor[HTML]{EFEFEF}0.0000 & \cellcolor[HTML]{EFEFEF}\checkmark \\
\multirow{-2}{*}{\textbf{Percep}}                                                     & G-Mean                                 & 0.0000                         & \checkmark                         & 0.0217                         & \checkmark                         & 0.0000                         & \checkmark                         & 0.0000                         & \checkmark                         \\
                                                                                      & \cellcolor[HTML]{EFEFEF}F1             & \cellcolor[HTML]{EFEFEF}0.0000 & \cellcolor[HTML]{EFEFEF}\checkmark & \cellcolor[HTML]{EFEFEF}0.0001 & \cellcolor[HTML]{EFEFEF}\checkmark & \cellcolor[HTML]{EFEFEF}0.0000 & \cellcolor[HTML]{EFEFEF}\checkmark & \cellcolor[HTML]{EFEFEF}0.0000 & \cellcolor[HTML]{EFEFEF}\checkmark \\
\multirow{-2}{*}{\textbf{MLP}}                                                        & G-Mean                                 & 0.0000                         & \checkmark                         & 0.0000                         & \checkmark                         & 0.0000                         & \checkmark                         & 0.0000                         & \checkmark                         \\
                                                                                      & \cellcolor[HTML]{EFEFEF}F1             & \cellcolor[HTML]{EFEFEF}0.2858 & \cellcolor[HTML]{EFEFEF}  & \cellcolor[HTML]{EFEFEF}0.5907 & \cellcolor[HTML]{EFEFEF}  & \cellcolor[HTML]{EFEFEF}0.4050 & \cellcolor[HTML]{EFEFEF}  & \cellcolor[HTML]{EFEFEF}0.5169 & \cellcolor[HTML]{EFEFEF}  \\
\multirow{-2}{*}{\textbf{RF}}                                                         & G-Mean                                 & 0.2229                         &                           & 0.5907                         &                           & 0.7535                         &                           & 0.4980                         &                           \\
                                                                                      & \cellcolor[HTML]{EFEFEF}F1             & \cellcolor[HTML]{EFEFEF}0.0933 & \cellcolor[HTML]{EFEFEF}  & \cellcolor[HTML]{EFEFEF}0.8063 & \cellcolor[HTML]{EFEFEF}  & \cellcolor[HTML]{EFEFEF}0.1639 & \cellcolor[HTML]{EFEFEF}  & \cellcolor[HTML]{EFEFEF}0.9574 & \cellcolor[HTML]{EFEFEF}  \\
\multirow{-2}{*}{\textbf{XGBoost}}                                                    & G-Mean                                 & 0.0933                         &                           & 0.8063                         &                           & 0.1639                         &                           & 0.9574                         &                           \\
                                                                                      & \cellcolor[HTML]{EFEFEF}F1             & \cellcolor[HTML]{EFEFEF}0.2448 & \cellcolor[HTML]{EFEFEF}  & \cellcolor[HTML]{EFEFEF}0.5662 & \cellcolor[HTML]{EFEFEF}  & \cellcolor[HTML]{EFEFEF}0.0137 & \cellcolor[HTML]{EFEFEF}\checkmark & \cellcolor[HTML]{EFEFEF}0.0705 & \cellcolor[HTML]{EFEFEF}  \\
\multirow{-2}{*}{\textbf{AdaBoost}}                                                   & G-Mean                                 & 0.2448                         &                           & 0.5662                         &                           & 0.0137                         & \checkmark                         & 0.0705                         &                           \\
                                                                                      & \cellcolor[HTML]{EFEFEF}F1             & \cellcolor[HTML]{EFEFEF}0.0041 & \cellcolor[HTML]{EFEFEF}\checkmark & \cellcolor[HTML]{EFEFEF}0.0078 & \cellcolor[HTML]{EFEFEF}\checkmark & \cellcolor[HTML]{EFEFEF}0.0000 & \cellcolor[HTML]{EFEFEF}\checkmark & \cellcolor[HTML]{EFEFEF}0.0000 & \cellcolor[HTML]{EFEFEF}\checkmark \\
\multirow{-2}{*}{\textbf{Bagging}}                                                    & G-Mean                                 & 0.0014                         & \checkmark                         & 0.0102                         & \checkmark                         & 0.0000                         & \checkmark                         & 0.0000                         & \checkmark                         \\
                                                                                      & \cellcolor[HTML]{EFEFEF}F1             & \cellcolor[HTML]{EFEFEF}0.5834 & \cellcolor[HTML]{EFEFEF}  & \cellcolor[HTML]{EFEFEF}0.6362 & \cellcolor[HTML]{EFEFEF}  & \cellcolor[HTML]{EFEFEF}0.0160 & \cellcolor[HTML]{EFEFEF}\checkmark & \cellcolor[HTML]{EFEFEF}0.0322 & \cellcolor[HTML]{EFEFEF}\checkmark \\
\multirow{-2}{*}{\textbf{OLA}}                                                        & G-Mean                                 & 0.6915                         &                           & 0.9364                         &                           & 0.0002                         & \checkmark                         & 0.0011                         & \checkmark                         \\
                                                                                      & \cellcolor[HTML]{EFEFEF}F1             & \cellcolor[HTML]{EFEFEF}0.0633 & \cellcolor[HTML]{EFEFEF}  & \cellcolor[HTML]{EFEFEF}0.0309 & \cellcolor[HTML]{EFEFEF}\checkmark & \cellcolor[HTML]{EFEFEF}0.0000 & \cellcolor[HTML]{EFEFEF}\checkmark & \cellcolor[HTML]{EFEFEF}0.0000 & \cellcolor[HTML]{EFEFEF}\checkmark \\
\multirow{-2}{*}{\textbf{LCA}}                                                        & G-Mean                                 & 0.0226                         & \checkmark                         & 0.0364                         & \checkmark                         & 0.0000                         & \checkmark                         & 0.0000                         & \checkmark                         \\
                                                                                      & \cellcolor[HTML]{EFEFEF}F1             & \cellcolor[HTML]{EFEFEF}0.1317 & \cellcolor[HTML]{EFEFEF}  & \cellcolor[HTML]{EFEFEF}0.8412 & \cellcolor[HTML]{EFEFEF}  & \cellcolor[HTML]{EFEFEF}0.3877 & \cellcolor[HTML]{EFEFEF}  & \cellcolor[HTML]{EFEFEF}0.0744 & \cellcolor[HTML]{EFEFEF}  \\
\multirow{-2}{*}{\textbf{MCB}}                                                        & G-Mean                                 & 0.1510                         &                           & 0.4385                         &                           & 0.1434                         &                           & 0.0256                         & \checkmark                         \\
                                                                                      & \cellcolor[HTML]{EFEFEF}F1             & \cellcolor[HTML]{EFEFEF}0.5368 & \cellcolor[HTML]{EFEFEF}  & \cellcolor[HTML]{EFEFEF}0.3985 & \cellcolor[HTML]{EFEFEF}  & \cellcolor[HTML]{EFEFEF}0.6109 & \cellcolor[HTML]{EFEFEF}  & \cellcolor[HTML]{EFEFEF}0.2935 & \cellcolor[HTML]{EFEFEF}  \\
\multirow{-2}{*}{\textbf{KNORAE}}                                                     & G-Mean                                 & 0.6592                         &                           & 0.2445                         &                           & 0.9499                         &                           & 0.7844                         &                           \\
                                                                                      & \cellcolor[HTML]{EFEFEF}F1             & \cellcolor[HTML]{EFEFEF}0.2378 & \cellcolor[HTML]{EFEFEF}  & \cellcolor[HTML]{EFEFEF}0.5939 & \cellcolor[HTML]{EFEFEF}  & \cellcolor[HTML]{EFEFEF}0.0000 & \cellcolor[HTML]{EFEFEF}\checkmark & \cellcolor[HTML]{EFEFEF}0.0000 & \cellcolor[HTML]{EFEFEF}\checkmark \\
\multirow{-2}{*}{\textbf{KNORAU}}                                                     & G-Mean                                 & 0.1257                         &                           & 0.3655                         &                           & 0.0000                         & \checkmark                         & 0.0000                         & \checkmark                         \\ \hline
                                                                                      & \cellcolor[HTML]{EFEFEF}F1             & \multicolumn{2}{c|}{\cellcolor[HTML]{EFEFEF}7}             & \multicolumn{2}{c|}{\cellcolor[HTML]{EFEFEF}5}             & \multicolumn{2}{c|}{\cellcolor[HTML]{EFEFEF}13}            & \multicolumn{2}{c}{\cellcolor[HTML]{EFEFEF}14}             \\
                                                                                      & G-Mean                                 & \multicolumn{2}{c|}{8}                                     & \multicolumn{2}{c|}{6}                                     & \multicolumn{2}{c|}{14}                                    & \multicolumn{2}{c}{14}                                     \\
\multirow{-3}{*}{\textbf{\begin{tabular}[c]{@{}l@{}}$H_0$\\ Rejections\end{tabular}}} & \cellcolor[HTML]{EFEFEF}\textbf{Total} & \multicolumn{2}{c|}{\cellcolor[HTML]{EFEFEF}\textbf{15}}   & \multicolumn{2}{c|}{\cellcolor[HTML]{EFEFEF}\textbf{11}}   & \multicolumn{2}{c|}{\cellcolor[HTML]{EFEFEF}\textbf{27}}   & \multicolumn{2}{c}{\cellcolor[HTML]{EFEFEF}\textbf{28}}   
\end{tabular}%
}
\end{table*}

%% file: tabs/tab_st_wins_per_ir_stratum.tex
\begin{table*}[ht]
\centering
\caption{Number of wins per scaling techniques considering all 82 models. Each IR stratum (low, medium and high) consider only the datasets within that stratum.  The best result for each row is highlighted in bold.}
\label{tab:st_wins_per_ir_stratum}
\resizebox{0.85\columnwidth}{!}{%
\begin{tabular}{lllllll}
\hline
\multicolumn{1}{c}{\textbf{IR Stratum}} & \multicolumn{1}{c}{\textbf{NS}} & \multicolumn{1}{c}{\textbf{SS}} & \multicolumn{1}{c}{\textbf{MM}} & \multicolumn{1}{c}{\textbf{MA}} & \multicolumn{1}{c}{\textbf{RS}} & \multicolumn{1}{c}{\textbf{QT}} \\ \hline
All                                     & 125.533                         & 168.100                         & 105.883                         & 109.467                         & 172.233                         & \textbf{220.783}                \\
Low IR                                  & 12.817                          & \textbf{32.567}                 & 15.733                          & 10.983                          & 23.450                          & 25.450                          \\
Medium IR                               & 15.900                          & 19.650                          & 11.983                          & 19.983                          & 22.983                          & \textbf{30.500}                 \\
High  IR                                & 96.817                          & 115.883                         & 78.167                          & 78.500                          & 125.800                         & \textbf{164.833}               
\end{tabular}%
}
\end{table*}

%% file: tabs/tab_related_works.tex
\begin{table*}[ht]
\centering
\caption{Amplitude of the related works and this paper.}
\label{tab:related_works}
\resizebox{1.0\textwidth}{!}{%
\begin{tabular}{c|l|c|l|l}
\multicolumn{1}{c|}{}                                           & \multicolumn{1}{c|}{}                                                                                                                                                                                                                                                                                                                                                 & \multicolumn{1}{c|}{}                                    & \multicolumn{2}{c}{\textbf{Classification Algorithms}}                                                                                                                                                                                                                                                                        \\
\multicolumn{1}{c|}{\multirow{-2}{*}{\textbf{Paper}}}           & \multicolumn{1}{c|}{\multirow{-2}{*}{\textbf{Scaling techniques}}}                                                                                                                                                                                                                                                                                                    & \multicolumn{1}{c|}{\multirow{-2}{*}{\textbf{Datasets}}} & \multicolumn{1}{c|}{Monolithic}                                                                                                                                                            & \multicolumn{1}{c}{Ensemble}                                                                                                      \\ \hline
\rowcolor[HTML]{EFEFEF} 
\begin{tabular}[c]{@{}c@{}}\textbf{Jain et al. \cite{jain2018}}\\2018\end{tabular}                            & \begin{tabular}[c]{@{}l@{}}\textbf{2 techniques:}\\ Min-max, Z-score.\end{tabular}                                                                                                                                                                                                                                                                                             & \textbf{48}                                              & \begin{tabular}[c]{@{}l@{}}\textbf{1 algorithm:} \\ Gaussian Kernel ELM.\end{tabular} & \begin{tabular}[c]{@{}l@{}}\end{tabular}                                                         \\
\begin{tabular}[c]{@{}c@{}}\textbf{Dzierżak et al. \cite{dzierzak2019}}\\2019\end{tabular}                    & \begin{tabular}[c]{@{}l@{}}\textbf{2 techniques:}\\ Min-max, Z-score.\end{tabular}                                                                                                                                                                                                                                                                                             & \textbf{1}                                               & \begin{tabular}[c]{@{}l@{}}\textbf{4 algorithms:}\\ Naive Bayes, SVM, \\ MLP, Classification \\ via regression.\end{tabular}                                                                        & \begin{tabular}[c]{@{}l@{}}\textbf{1 algorithm:}\\ Random Forests (RF)\end{tabular}                                                         \\
\rowcolor[HTML]{EFEFEF} 
\begin{tabular}[c]{@{}c@{}}\textbf{Raju et al. \cite{raju2020}}\\2020\end{tabular}                           & \begin{tabular}[c]{@{}l@{}}\textbf{7 techniques:}\\ Min-max, Z-score, Scale, \\ Robust scaler, Quantile \\ Transform, Power \\ Transform,  Max abs scaler.\end{tabular}                                                                                                                                                                                                            & \textbf{1}                                               & \begin{tabular}[c]{@{}l@{}}\textbf{3 algorithms:}  \\ SVM (2 variants), KNN.\end{tabular}                                                                                                           &                                                                                                                                   \\
\cellcolor[HTML]{FFFFFF}
\begin{tabular}[c]{@{}c@{}}\textbf{Singh et al. \cite{singh2020a}}\\2020\end{tabular} & \cellcolor[HTML]{FFFFFF}\begin{tabular}[c]{@{}l@{}}\textbf{14 techniques:}\\ Z-score, Min-max (2 variants), \\ Mean centering, Pareto scaling, \\ Variable stability scaling, \\ Power Transform, Max abs \\ scaler, Decimal scaling, \\ Median and Median Abs \\ Deviation Normalization, \\ Tanh normalization (2 variants),\\ Logistic sigmoid, Hiperb. tangent.\end{tabular} & \cellcolor[HTML]{FFFFFF}\textbf{21}                      & \cellcolor[HTML]{FFFFFF}\textbf{1 algorithm:} KNN                                                                                                                                                   &                                                                                                                                   \\
\rowcolor[HTML]{EFEFEF} 
\begin{tabular}[c]{@{}c@{}}\textbf{Mishkov et al. \cite{Mishkov2022}}\\2022\end{tabular}                     & \begin{tabular}[c]{@{}l@{}}\textbf{16 techniques:} \\ Z-score, Pos. standardization, \\ Unitization, Min-max, \\ Normalization (3 variants), \\ Pos. normalization (2 variants), \\ Quotient transform. (7 variants).\end{tabular}                                                                                                                                & \textbf{4}                                               & \textbf{1 algorithm:} MLP                                                                                                                                                                           &                                                                                                                                   \\ \hline
\textbf{This paper}                                             & \begin{tabular}[c]{@{}l@{}}\textbf{5 techniques:}\\ Min-max, Z-score,\\ Max abs scaler, \\ Robust Scaler,\\Quantile Transformer.\end{tabular}                                                                                                                                                                                                                                     & \textbf{82}                                              & \begin{tabular}[c]{@{}l@{}}\textbf{11 algorithms:} \\ SVM (2 variants), KNN, \\ GLVQ, GNB, GP, LDA, \\ QDA, DT, Percep, MLP.\end{tabular}                                                           & \begin{tabular}[c]{@{}l@{}}\textbf{9 algorithms:}\\ XGBoost,  RF, \\ AdaBoost, Bagging, \\ OLA, LCA, MCB, \\ KNORAE,  KNORAU.\end{tabular}
\end{tabular}%
}
\end{table*}